\documentclass{ecai} 

\usepackage{latexsym}
\usepackage{amssymb}
\usepackage{amsmath}
\usepackage{amsthm}
\usepackage{booktabs}
\usepackage{enumitem}
\usepackage{graphicx}
\usepackage{color}
\usepackage{microtype}

\usepackage{algorithm}
\usepackage{algorithmic}
\usepackage[switch]{lineno}
\usepackage{multirow}
\usepackage{subfig}
\usepackage{tabularx}
\newtheorem{theorem}{Theorem}
\newtheorem{lemma}[theorem]{Lemma}
\newtheorem{corollary}[theorem]{Corollary}
\newtheorem{proposition}[theorem]{Proposition}

\newtheorem{definition}{Definition}

\newcommand{\BibTeX}{B\kern-.05em{\sc i\kern-.025em b}\kern-.08em\TeX}

\begin{document}


\begin{frontmatter}




\title{Graph Structure Learning with Temporal Graph Information Bottleneck for Inductive Representation Learning}


\author[A]{\fnms{Jiafeng}~\snm{Xiong}\orcid{0000-0002-3789-3063}\thanks{Corresponding Author. Email: jiafeng.xiong@manchester.ac.uk.}}
\author[A]{\fnms{Rizos}~\snm{Sakellariou}\orcid{0000-0002-6104-6649}} 

\address[A]{Department of Computer Science, University of Manchester, UK}


\begin{abstract}
Temporal graph learning is crucial for dynamic networks where nodes and edges evolve over time and new nodes continuously join the system. Inductive representation learning in such settings faces two major challenges: effectively representing unseen nodes and mitigating noisy or redundant graph information. We propose \textbf{GTGIB}, a versatile framework that integrates \textbf{\underline{G}}raph Structure Learning (GSL) with \textbf{\underline{T}}emporal \textbf{\underline{G}}raph \textbf{\underline{I}}nformation \textbf{\underline{B}}ottleneck (TGIB). We design a novel two-step GSL-based structural enhancer to enrich and optimize node neighborhoods and demonstrate its effectiveness and efficiency through theoretical proofs and experiments. The TGIB refines the optimized graph by extending the information bottleneck principle to temporal graphs, regularizing both edges and features based on our derived tractable TGIB objective function via variational approximation, enabling stable and efficient optimization. GTGIB-based models are evaluated to predict links on four real-world datasets; they outperform existing methods in all datasets under the inductive setting, with significant and consistent improvement in the transductive setting.
\end{abstract}

\end{frontmatter}


\section{Introduction}
Recent years have witnessed a significant surge of interest in temporal graph representation learning \cite{cong_we_2023,gravina_long_2024,tian_freedyg_2024}. In many real-world scenarios such as social networks, communication networks and recommendation systems \cite{skarding_foundations_2021,jiang_graph-based_2022,wu_graph_2022}, the underlying graphs are continuously evolving, making temporal networks increasingly dynamic with continuous node emergence~\cite{holme_temporal_2012}. Despite the rapid advances, one essential issue is often overlooked: the inductive capability of these temporal graph representations. Inductive representation learning aims to generate embeddings and make predictions for newly emerged, unseen nodes~\cite{trivedi_dyrep_2019}. This capability is critical for dynamic networks where new entities constantly join the system, e.g., new users in social media~\cite{kovanen_temporal_2013}. Hence, the temporal graph learning model must learn and generalize the underlying dynamic patterns.

However, current methods for temporal graph methods face two major challenges when addressing inductive representation learning. Firstly, some existing approaches fail to effectively represent unseen nodes—methods that rely on node identities are inherently non-inductive~\cite{goyal_dyngem_2018,chen_e-lstm-d_2021}, while many structure-based methods struggle with new nodes that lack historical neighbors because they rely on past neighbor information to build node representations~\cite{xu_inductive_2020,wang_inductive_2022}. Secondly, real-world temporal networks often contain noise and redundant information and many current methods are insufficient in filtering this information, potentially leading to performance limitations~\cite{battaglia_relational_2018}.  These problems impair inductive representation and generalization for downstream tasks, yet few existing methods explicitly address them. 

Building on the Information Bottleneck (IB) principle~\cite{tishby_information_2000}, which learns representations that are mutually maximally predictive yet minimally informative of the input, graph‐specific IB extensions have effectively filtered noise for robust representations~\cite{wu_graph_2020,yang_heterogeneous_2021} and DGIB~\cite{yuan_dynamic_2024} extends IB to dynamic graphs. But these methods are limited to static or discrete‐time snapshots with global adjacency matrices, undermining temporal fidelity and precluding inductive representation learning.

In this paper, we propose a versatile framework, GTGIB, integrating \textbf{\underline{G}}raph Structure Learning (GSL)~\cite{jin_graph_2020,liu_towards_2022} with \textbf{\underline{T}}emporal \textbf{\underline{G}}raph \textbf{\underline{IB}} (TGIB). Specifically, to solve the insufficient representation of unseen nodes, we introduce a novel structure enhancer based on GSL. This module samples and learns to construct underlying edges to enrich the temporal graph neighbors to facilitate inductive representation learning. To address the redundancy and bias from the dataset and structure enhancer, we extend the IB into continuous-time dynamic graphs (CTDG), which regularize the graph structures and node features and represent more dynamics than discrete-time snapshots~\cite{souza_provably_2022}. TGIB effectively filters out noisy or irrelevant links, maintaining temporal dynamics, leading to more succinct and task-relevant representations. Finally, we employ the TGN~\cite{rossi_temporal_2020} and CAW~\cite{wang_inductive_2022} models with GTGIB because of their distinct graph learning paradigms, message-passing and random-walk mechanisms, respectively. We utilize reparameterization for temporal link prediction (TLP) on the refined graph.
\noindent\textbf{Contributions}:

(i) A novel IB-based framework based on CTDG, GTGIB, with a new two-step GSL-based structure enhancer is proposed. GTGIB is general and not tied to specific backbones, as it applies to data before graph representation. Theoretical and experimental analysis also confirm that the structure enhancer is effective and efficient.

(ii) A tractable variational upper bound for TGIB is derived, based on the continuous-time Markov chain and CTDG. By incorporating time-dependent Bernoulli distribution and reparameterization, TGIB jointly regularizes temporal structures and node features, mitigating irrelevant information in representation learning.

(iii) Experimental evaluation on four real-world TLP datasets shows that GTGIB flexibly supports different representation paradigms and consistently boosts TGN and CAW performance. It outperforms existing inductive methods, improving base models by an average of 3.03\% for TGN and 3.17\% for CAW.

\section{Related Work}
\subsection{Temporal Graph Representation Learning}

Temporal networks can be represented as discrete-time dynamic graphs (DTDG) or continuous-time dynamic graphs (CTDG). DTDGs consist of a series of graph snapshots with fixed time intervals. Early works~\cite{chen_e-lstm-d_2021,goyal_dyngem_2018} apply Graph Neural Networks, such as GCN~\cite{kipf_semi-supervised_2017} or GAT~\cite{velickovic_graph_2018}, to these snapshots and aggregate the outputs for downstream tasks~\cite{xu_inductive_2020,wang_inductive_2022}. However, these methods fail to capture irregular event arrivals, limiting their effectiveness in dynamic environments.

CTDGs model temporal networks as sequences of timestamped events occurring at irregular times. Methods such as DyRep~\cite{trivedi_dyrep_2019} and JODIE~\cite{kumar_predicting_2019} utilize Recurrent Neural Networks (RNNs) to update node embeddings. In contrast, TGAT~\cite{xu_inductive_2020}, TGN~\cite{rossi_temporal_2020} and PINT~\cite{souza_provably_2022} employ message passing (MP) for neighborhood aggregation. CTDNE~\cite{nguyen_continuous-time_2018} and CAW~\cite{wang_inductive_2022} leverage random walks (RW) to capture local graph structures. Furthermore, other temporal graph learning methods adopt diverse approaches, such as neighbor pooling, Transformer architectures and node frequency, as demonstrated by GraphMixer~\cite{cong_we_2023}, DyGFormer~\cite{yu_towards_2023} and FreeDyG~\cite{tian_freedyg_2024}, respectively. Despite these advancements, existing methods encounter significant limitations in inductive learning scenarios, as representations for unseen nodes rely on node identities or lack sufficient historical neighbors.

\subsection{Information Bottleneck on Graphs}

The concept of IB~\cite{tishby_information_2000} has gained considerable attention in deep learning research due to its ability to balance representation learning and information compression~\cite{alemi_deep_2019}. In the context of graph data, several works have explored the application of IB principles. GIB~\cite{wu_graph_2020} extends IB to graph-structured data by jointly regularizing structural and feature information to learn robust node representations. For specific graph tasks, SIB~\cite{yu_graph_2020} addresses subgraph recognition through IB mechanisms, while HGIB~\cite{yang_heterogeneous_2021} leverages IB principles to learn representations in static heterogeneous networks. TempME~\cite{chen_tempme_2024} employs IB to assist in identifying temporal motifs, however, it does not consider temporal constraints nor the IB of graph features. Besides, DGIB~\cite {yuan_dynamic_2024} based on DTDG, compels it to use the Consensual Constraint to simplify temporal information to a fixed ordering of snapshot sequence—an inherent loss of temporal fidelity that CTDG avoid~\cite{souza_provably_2022}. DGIB’s dependence on the global snapshot adjacency matrix limits it to only supporting transductive prediction. Overall, the IB framework for temporal graph representation learning has not been adequately explored and there is scope for further research.

\subsection{Graph Structure Learning}
 
Given the inherent limitations of predefined graph structures, GSL has emerged as a promising approach to optimizing and refining network topology. GSL methods utilize a structured learner to compute edge weights for refining the graph, that is, removing non-essential edges~\cite{jin_graph_2020} or to complete the graph through sampling techniques~\cite{zhang_time-aware_2023,marisca_graph-based_2024}. This task-guided refinement uncovers more informative relationships that better capture the latent dependencies between nodes. Importantly, GSL is conceptually distinct from graph augmentation. Graph augmentation typically perturbs an existing graph during preprocessing or training to boost data diversity and robustness without learning a new structure~\cite{adjeisah_towards_2023}. In contrast, GSL optimizes graph topology to meet the downstream task’s objectives. 

The development of GSL approaches has followed three main directions. Early metric-based methods relied on non-learnable similarity functions like Gaussian kernels to determine edge weights~\cite{yu_graph-revised_2020}, offering computational efficiency but limited expressiveness. More recent neural-based approaches leverage deep models to learn complex edge relationships~\cite{liu_towards_2022,wei_contrastive_2022}, achieving superior performance through their ability to capture non-linear patterns. Direct optimization methods treat the adjacency matrix as learnable parameters~\cite{jin_graph_2020}, providing flexibility at the cost of increased optimization complexity.

Significant advances have been made in incorporating IB into GSL. These approaches leverage principles such as mutual information maximization and IB theory to optimize graph structures. For instance, VIB-GSL~\cite{sun_graph_2022} utilizes variational IB to retain task-relevant information. Similarly, CGI~\cite{wei_contrastive_2022} combines the IB principle with contrastive learning to optimize graph structures in recommender systems. These information-theoretic strategies enhance the robustness and expressiveness of GSL. However, these approaches primarily target static graphs and overlook temporal dynamics.

\section{Preliminaries}
\subsection{Notations and Problem Definition} 

In a CTDG temporal graph ${G}(t) = (V(t), E(t), \mathcal{X},\mathcal{E})$ with the time domain $T=\{t_1,t_2,\dots\mid t_1<t_2<\dots\}$, $V(t) \subseteq V$ is the set of nodes active at time $t$, where $V = \{1, 2, \ldots, n\}$ is the set of all nodes and $ E(t) \subseteq \{(i, j, t') | i, j \in V(t), t' \leq t\}$ means a sequence of timestamped events w.r.t. edge, between nodes $i, j$ at timestamp $t'$. Each event may have an edge feature $e_{ij}(t) \in \mathcal{E}$ and $ \mathcal{E} \in \mathbb{R}^k$ is the edge feature set. Each node $i$ has an unchanged initial node feature ${x}_i \in \mathcal{X}$. $\mathcal{X} \in \mathbb{R}^d$ is the node feature set. Additionally, the temporal neighborhood $\mathcal{N}_i(t)$ of the node $i$ is defined as $\mathcal{N}_i(t) = \{ (j, t') \mid \exists (j, i, t') \in G(t), t' < t \}$. We can also recursively define $l$-hop temporal neighbors $\mathcal{N}^{(l)}_i(t) = \{ (k, t') \mid \exists (j, t'') \in \mathcal{N}^{(l-1)}_i(t), (k, j, t') \in E(t''), t' < t'' \}$. Besides, $ G([t - \Delta t, t]) $ represents the temporal graphs from the timestamp $ t - \Delta t $ to $ t $.

Temporal link prediction comprises transductive or inductive settings~\cite{xiong_survey_2025}. In the transductive setting, the model predicts links among nodes accessed during training. In contrast, the inductive setting requires the model to generalize learned temporal patterns to predict links involving nodes unseen during training.

\subsection{Information Bottleneck}
The IB principle aims to find an optimal encoding $Z$ of input data $X$ that retains maximal information about the target $Y$ while being maximally compressed. The IB framework utilizes several concepts from information theory, including Shannon entropy $H(X)$, mutual information $I(X; Y)$  and Kullback-Leibler divergence $\mathcal{D}_{KL}(\cdot\parallel\cdot)$. These concepts provide the theoretical foundation for our temporal graph structure learning approach.

\begin{definition}[Information Bottleneck Principle]
Given input data $X$, its label $Y$ and representation $Z$ following the Markov Chain $\langle Y \rightarrow X \rightarrow Z \rangle$, the optimal representation $Z$ is obtained by:
\begin{equation}
Z = \arg\min_Z -I(Z; Y) + \gamma I(Z; X),
\end{equation}
where $\gamma$ is Lagrange multiplier regulating information compression.
\end{definition}

The variational IB objective~\cite{alemi_deep_2019}, parameterized by a neural network:
\begin{equation}
\begin{split}
\mathcal{L}_{\text{VIB}} = \frac{1}{N} \sum_{i=1}^{N} \int dZ p(Z\mid X_i) \log q(Y_i\mid Z)  +\\ 
+ \gamma \mathcal{D}_{KL}(p(Z\mid X_i)\parallel q(Z)),
\end{split}
\end{equation}
where $X_i$ and $Y_i$ denote the $i$-th $(1 \leq i \leq N)$ input sample and its label, $q(\cdot)$ is the variational approximation to the probabilistic distribution functions (PDF) $p(\cdot)$. 

\subsection{Temporal Graph Networks}
\begin{figure}[t]
    \centering
    \includegraphics[width=.72\linewidth]{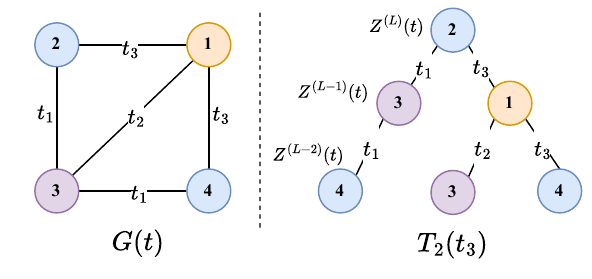}
    \caption{An example of a temporal graph and a temporal computation tree rooted at node 2, where the timestamps $t_1 < t_2 < t_3$.}
    \label{fig:TCT}
\end{figure}
Temporal Graph Networks (TGNs) \cite{rossi_temporal_2020} is a general temporal graph learning framework with three key components: aggregation, update and memory. In this paper, we denote $ L $ as the number of TGN layers. TGNs compute the node embedding $ Z^{(l)}_i(t) $ of node $ i $ at layer $ l $ ($ 1 \leq l \leq L $) at current timestamp $t$ by recursively applying:
\begin{equation}
\tilde{Z}^{(l)}_i(t) = \text{AGG}^{(l)}\bigl(Z^{(l-1)}_j(t), t - t', e_{ij}(t')\bigr),
\label{eq:agg}
\end{equation}
\begin{equation}
Z^{(l)}_i(t) = \text{UPDATE}^{(l)}\bigl(Z^{(l-1)}_i(t), \tilde{Z}^{(l)}_i(t)\bigr),
\label{eq:update}
\end{equation}
where $\text{AGG}^{(l)}$ and $\text{UPDATE}^{(l)}$ are parameterized functions, $(j, t')$ is the temporal neighborhood $\mathcal{N}_i(t)$ and $e_{ij}(t')$ is the edge feature of the edge $(i,j,t')$. The memory consists of vectors that summarize the node's historical embedding and are updated as events occur. 

The message-passing computations for node $i$ at timestamp $t$ are represented by its temporal computation tree (TCT) $T_i(t)$. This tree has node $i$ as its root and height $L$. The leaves at level $l$ correspond to the message-passing operations at layer $l$ in the TGN. As shown in Figure~\ref{fig:TCT}, the node embedding $Z^{(L)}(t)$ at level $L$ is obtained by aggregating and updating the embeddings $Z^{(L-1)}(t)$ from its temporal neighborhood leaves at level $L-1$.

\subsection{Continuous-Time Markov Chains}

A continuous-time Markov chain (CTMC)~\cite{anderson_continuous-time_1991} is a stochastic process $\{X(t) : t \geq 0\}$ defined on a countable state space $\Omega$. It satisfies the Markov property, meaning that for any timestamp $t, \Delta t \geq 0$ and $i, j \in \Omega$:
\begin{align}
P(X(t) = j \mid & X(t - \Delta t) = i, \{X(t') : t' < t - \Delta t\}) \nonumber \\
&
= P(X(t) = j \mid X(t - \Delta t) = i).
\end{align}
In temporal networks, a CTMC assumes node or edge transitions depend only on the state at \(t-\Delta t\). \(\Delta t\) is the observation window.

\section{Methodology}
\subsection{Overview of Framework}
\begin{figure*}
    \centering
    \includegraphics[width=1\linewidth]{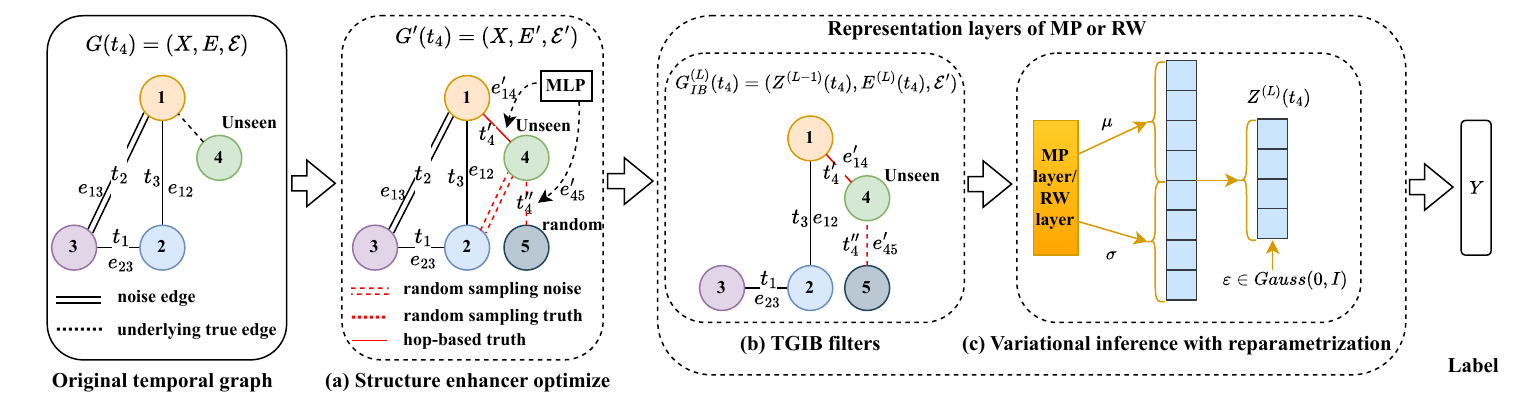}
    \caption{Overview of the framework. The process includes (a) a temporal graph optimized by GSL-based structure enhancer to sample candidate edges and generate edge features and timestamps, (b) filtering with TGIB module to remove noise and irrelevant or redundant information and (c) variational inference for temporal link prediction (TLP). Node 4 represents an unseen node without neighbors, appearing at timestamp $t_4$.}
    \label{fig:framework}
\end{figure*}
Our framework consists of three main components: (a) a GSL-based structure enhancer for temporal graph optimization, (b) a TGIB module for filtering and (c) a variational inference module for TLP.

First, Figure~\ref{fig:framework}(a) illustrates the graph structure learning performed by the two-step structure enhancer on the original temporal graph $G(t_4)$ at timestamp $t_4$. In this enhancement process, candidate edges are first constructed through complementary global random sampling and locally hop-based sampling, and then a multi-layer perceptron (MLP) generates edge features to transform $G(t_4)$ into a new graph $G'(t_4)$, where both the noisy edges and the underlying temporal edges are retained while additional potential temporal neighbors are added.

Second, in Figure~\ref{fig:framework}(b), the TGIB filters the noise and generates the IB graph $G^{(L)}_{\text{IB}}(t_4)$ based on $G'(t_4)$ and the node embeddings  $Z^{(L-1)}(t_4)$ from the previous layer. The TGIB is processed before the message passing or random walk to refine the graph structure. 

Finally, in Figure~\ref{fig:framework}(c), the refined graph undergoes variational inference through a temporal graph representation model such as TGN or CAW to learn the distribution of embeddings. The distribution is then utilized to generate the node embeddings $Z^{(L)}(t_4)$ via reparameterization. This design ensures our framework remains flexible and adaptable to various temporal graph learning architectures.

\subsection{Structure Enhancer}

\begin{figure}
    \centering
    \includegraphics[width=1\linewidth]{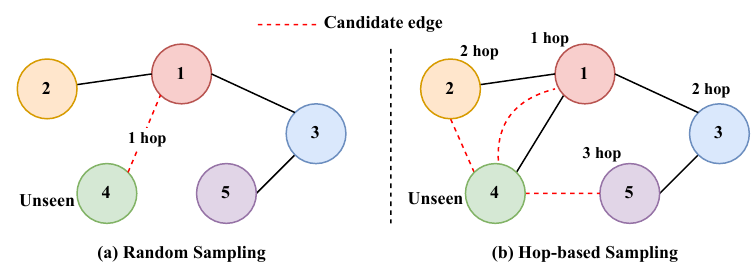}
    \caption{Node 4 is unseen. The enhancer first uses (a) random sampling to add only 1-hop neighbors, Node 1, globally, then uses (b) hop-based sampling to expand multi-hop neighbors.}
    \label{fig:enhancer}
\end{figure}

Current Graph Structure Learning (GSL) methods for temporal graphs directly learn edge weights between  all node pairs to optimize topology, which is memory-intensive and computationally inefficient. The key is to have efficiency and diversity while maintaining the learning of edge weights effectively. Thus, we propose a two-step GSL-based structure enhancer. 

In the first step, the structure enhancer uses a complementary sampling strategy to construct candidate edges, as illustrated in Figure~\ref{fig:enhancer}. To explore diverse relationships efficiently, a random sampling strategy is first adopted to select destination nodes globally. This approach helps discover potential long-range connections and introduces diverse structural patterns that local-only methods might overlook. Then, a flexible hop-based sampling strategy is designed to sample multiple destination nodes at different hop distances locally.  For an unseen node without any neighbors, random sampling firstly establishes initial connections, after which the hop-based strategy is employed to enrich the neighbors. 

In the second step, for each newly generated edge, a timestamp $t_{new}$ is randomly sampled from the range $[0,t]$, where $t$ denotes the current timestamp. The edge features are then computed as $\text{MLP}(x_i \oplus x_j \oplus \text{TE}(t - t_{new}))$, where $x_i$ and $x_j$ are node features, $\oplus$ denotes the concatenation operation, $\text{MLP}$ is a multi-layer perceptron and $\text{TE}(\cdot)$ represents the generic time encoding~\cite{xu_inductive_2020}. Our TGIB and task objectives will continuously optimize the new graph structure. The proofs and time analysis in the Appendix, Sections~\ref{appendix:structure enhancer} and~\ref{appendix:train time} show the effectiveness and efficiency of the structure enhancer.

\subsection{Deriving the Temporal Graph Information Bottleneck Principles}

In general, the IB principle relies on the assumption of independent and identically distributed data. However, the interconnected nature of graph-structured data violates this assumption. Therefore, most GIB-based models typically rely on a widely accepted local-dependence assumption~\cite{wu_graph_2020} for graph-structured data: given the data related to the neighbors within a certain number of hops of a node $i$, the data in the rest of the graph is independent of the node $i$. 

\paragraph{Temporal local dependence assumption.}

In temporal settings, the embedding of  the node $ i $ at timestamp $ t $, $ Z_i(t) $, depends only on its previous embedding $ Z_i(t - \Delta t) $ and the embeddings of its $l$-hop temporal neighbors $ \{Z_j(t) \mid j \in \mathcal{N}^{(l)}_i(t)\} $, independent of the rest of the graph. Formally, under the CTMC framework:
\begin{align}
    P(&Z_i(t) \mid  Z_i(t - \Delta t), 
    \{Z_j(t) \mid j \in \mathcal{N}^{(l)}_i(t)\},
    G(t - \Delta t))) = \nonumber \\
    &= P(Z_i(t) \mid Z_i(t - \Delta t), 
    \{Z_j(t) \mid j \in \mathcal{N}^{(l)}_i(t)\}).
\end{align}
This indicates that $ Z_i(t) $ evolves based only on its previous state and $l$-hop temporal neighbors, independent of $ G(t - \Delta t) $. For instance, in Figure~\ref{fig:TCT}, for TCT $T_2(t_3)$ with an effective observation window $\Delta t \le t_3-t_2$, node $2$ depends solely on node $1$ at $t_3$, independent of node $3$ and all $(L-2)$-layer neighbors of both nodes.

For an $L$-layer model, predictions are derived from the final layer embeddings $Z^{(L)}(t)$ at timestamp $t$. Thus, based on the temporal local dependence assumption, the TGIB objective simplifies as:
\begin{equation}
    \text{TGIB} \triangleq \bigl[ -I( Z^{(L)}(t);Y) + \gamma I(Z^{(L)}(t);G([t-\Delta t,t])) \bigr].
    \label{eq:tgib}
\end{equation}
The TGIB objective in \eqref{eq:tgib} consists of two terms: $-I(Z^{(L)}(t); Y)$, which maximizes the mutual information between the learned representation $Z^{(L)}(t)$ and the target $Y$, thereby encouraging predictive power; and $\gamma I(Z^{(L)}(t); G([t-\Delta t,t]))$, which regularizes the information in $Z^{(L)}(t)$ about the related history $G([t-\Delta t,t])$, thus promoting the compression of noise. The hyperparameter $\gamma$ controls the degree of compression.

As a foundation, we clarify that the task-irrelevant information introduced by the structure enhancer does not affect our IB graph $G_{\text{IB}}(t)$ learning from TGIB by the following lemma.
\begin{lemma}[Nuisance Invariance]\label{lemma:nuisance}
Given a temporal graph $G(t)$ associated with a label $Y$. Let $G_n$  be the task-irrelevant nuisance for $Y$. Denote ${G}_{{\text{IB}}}(t)$ as the refined graph based on IB objective. Then:
\begin{align}
I\bigl(G_{\text{IB}}(t); G_n\bigr)
\le
I\bigl(G_{\text{IB}}(t); G(t)\bigr)
-
I\bigl(G_{\text{IB}}(t); Y\bigr)
\label{eq:lemma}
\end{align}
\end{lemma}
Please refer to the Appendix, Section~\ref{appendix:lemma} for a detailed proof.

Lemma~\ref{lemma:nuisance} demonstrates that, under the IB objective, if the structure enhancer introduces irrelevant edges, the refined IB graph’s dependence on extraneous information does not increase. 

Second, to ensure TGIB tractability, we analyze both the prediction term $-I\bigl(Z^{(L)}(t);Y \bigr)$ and regularization term $I\bigl(Z^{(L)}(t);G([t-\Delta t,t])\bigr)$ in \eqref{eq:tgib}, with upper bounds given by Proposition~\ref{prop:upper_bounds}.
\begin{proposition}[Upper Bounds of TGIB]\label{prop:upper_bounds}
\begin{align}
&-I\bigl(Y;Z^{(L)}(t)\bigr)
\le 
-\frac{1}{N} \sum_{i=1}^{N} \log q\bigl(Y_i \mid Z^{(L)}_i(t)\bigr),
\label{eq:upper_bound1} \\
&I\bigl(Z^{(L)}(t);G([t-\Delta t,t])\bigr) 
\le
\sum_{l=1}^{L} (\mathrm{EIB}^{(l)}+\mathrm{XIB}^{(l)}),
\label{eq:upper_bound2}
\end{align}
where $\mathrm{XIB}^{(l)}$ and $\mathrm{EIB}^{(l)}$ are defined as follows:
\begin{align}
\mathrm{XIB}^{(l)} 
= \mathbb{E} \Bigl[\log \frac{p\bigl(Z^{(l)}(t) \mid G([t-\Delta t,t])\bigr)}{q\bigl(Z^{(l)}(t)\bigr)}\Bigr],
\label{eq:xib}\\
\mathrm{EIB}^{(l)} 
= \mathbb{E} \Bigl[\log \frac{p\bigl(E^{(l)}(t) \mid G([t-\Delta t,t])\bigr)}{q\bigl(E^{(l)}(t)\bigr)}\Bigr],
\label{eq:eib}
\end{align}
with $\mathbb{E}$ denoting expectation,$Y_i$ as the corresponding label, $Z^{(L)}_i(t)$ representing the node embedding generated based on the relevant temporal neighborhood at layer $L$, $ Z^{(l)}(t) $ and $ E^{(l)}(t), (1\leq l \leq L)$ are hidden refined embedding and graph structure at layer $ l $ and $q\bigl(Y_i \mid Z^{(L)}_i(t)\bigr), q\bigl(E^{(l)}(t)\bigr)$ and $q\bigl(Z^{(l)}(t)\bigr)$ being the variational approximation of their corresponding true posterior. 
\end{proposition}

Introducing the variational distributions $q(Y_i|Z^{(L)}_i(t))$, $q(Z^{(l)}(t))$ and $q(E^{(l)}(t))$ is necessary because directly optimizing the true posterior distributions is challenging. The variational distributions $q(Z^{(l)}(t))$ and $q(E^{(l)}(t))$ facilitate the tractability of the upper bounds. The proofs are in the Appendix, Section~\ref{appendix:upper bounds}. XIB and EIB regularize node features and graph structure, respectively.

Finally, to allow more flexibility (in a similar spirit as $\beta$-VAE~\cite{higgins_beta-vae_2017}), we allow the original coefficient $\gamma$ before EIB and XIB to be different and denote them as $\alpha$ and $\beta$. In summary, the objective is written as:
\begin{equation}
\text{TGIB}
\le
-\frac{1}{N} \sum_{i=1}^{N} \log q(Y_i| Z^{(L)}_i(t)) 
+
 \sum_{l=1}^{L} (\alpha  \text{EIB}^{(l)} 
+
 \beta \text{XIB}^{(l)}).
\label{eq:final bound}
\end{equation}

\subsection{Instantiating the TGIB Principle}\label{sec:instantiating}
\begin{algorithm}[htb]
\caption{The overall process of GTGIB-TGN}
\label{alg:tgib}
\textbf{Input}: Temporal Graph $G(t)=(V(t), E(t), \mathcal{X}, \mathcal{E})$; time domain $T$; node $i, j \in V(t)$.\\
\textbf{Initialize}: $Z^{(0)}(t)  \leftarrow \mathcal X$; $\mu_i^{(l)}, \sigma_i^{(l)} \in \mathbb{R}^{K \times 1}$; $\varepsilon \sim \text{Gauss}(0,I)$;$W \in \mathbb{R}^{2K \times 1}$.\\
\textbf{Output}: ${Z^{(L)}(t)}$, $\hat{Y}$.
\begin{algorithmic}[1]
\FOR{current time $t \in T$}
\STATE $Y_{ij}^{\text{pos}} \sim E(t), Y_{ij'}^{\text{neg}} \sim \text{NegativeSampling}(E(t))$ 
\STATE // Structure enhance
\STATE $G'(t)\leftarrow \text{Random Sampling Enhance}(i, j,j',G(t))$
\STATE $G'(t)\leftarrow \text{Hop-based Sampling Enhance}(i,j,j',G'(t))$
\STATE $G'(t)=(V(t), E'(t), \mathcal{X}, \mathcal{E}')\leftarrow \text{MLP Edge Features and Timestamps Generations}(i,j,j',G'(t))$
\FOR {layers $l = 1, 2, \cdots, L$}
\STATE // Temporal neighborhood filtering for $G'(t)$
\STATE $E^{(l)}_{\text{IB}}(t) \leftarrow \bigcup_{(i,j,t') \in E'(t)}{\varphi_{ij} \sim \text{Ber}(\pi_{ij}^{(l)}(t'))}$
\STATE $G_{\text{IB}}^{(l)}(t) \leftarrow (E^{(l)}_{\text{IB}}(t), Z^{(l-1)}(t))$
\STATE // Learn distribution
\STATE $[\mu^{(l)} \oplus \log \sigma^{(l)}] \leftarrow {\text{AGG}}^{(l)}_{\text{TGN}}(G_{\text{IB}}^{(l)}(t))$
\STATE // Sample graph representation
\STATE Reparameterize $Z^{(l)}(t) = \mu^{(l)} + \sigma^{(l)} \odot \varepsilon$
\ENDFOR
\STATE // Optimize
\STATE $\hat{Y}_{ij}^{\text{pos}} = \text{sigmoid}\bigl([Z_{i}^{(L)}(t) \oplus Z_{j}^{(L)}(t)] W\bigr), \hat{Y}_{ij'}^{\text{neg}} = \text{sigmoid}\bigl([Z_{i}^{(L)}(t) \oplus Z_{j'}^{(L)}(t)] W\bigr)$
\STATE $\mathcal{L}_{\text{TGIB}} = \mathcal{L}_{\text{CE}}(\hat{Y}, Y) +  \sum^{L}_{l=1}(\alpha\widehat{\text{EIB}}^{(l)}+\beta \widehat{\text{XIB}}^{(l)})$
\ENDFOR
\end{algorithmic}
\end{algorithm}
The GTGIB is versatile and can be integrated into different models. We demonstrate an application algorithm example by incorporating it with TGN models that utilize multi-head attention aggregation~\cite{vaswani_attention_2017}, resulting in the GTGIB-TGN model presented in Algorithm~\ref{alg:tgib}.  It details the GTGIB-TGN for TLP. The negative sampling in the algorithm selects edges absent during the target interval as negative training examples. Furthermore, we explore a CAW variant of GTGIB-CAW and evaluate its performance in Section~\ref{sec:experiments}.

\textbf{(1) Temporal Neighborhood Filtering.}
Within the GTGIB framework, we filter the graph structure at a given time $ t $ to eliminate noisy edges while retaining informative connections once the structure enhancer optimizes the graph. Specifically, for each edge $(i,j,t') \in E(t)$, we model its retention as a concrete relaxation Bernoulli~\cite{jang_categorical_2017} random variable $\varphi_{ij} \sim \text{Ber}(\pi_{ij}(t_{ij}))$:
\begin{equation}
    \text{Ber}(\pi_{ij}(t_{ij})) \approx \text{sigmoid}\bigr(\frac{1}{\tau}(\log \frac{\pi_{ij}(t_{ij})}{1-\pi_{ij}(t_{ij})} + \log \frac{\epsilon}{1-\epsilon})\bigl),
\end{equation}
\begin{equation}
    \pi_{ij}(t_{ij}) = \text{MLP}\left(\left[ x_i \oplus x_j \oplus \text{TE}(t-t_{ij}) \oplus e_{ij}(t_{ij}) \right]\right),
\end{equation}
where $\epsilon \sim \text{Uniform}(0, 1)$; $\tau \in (0,1)$  is the temperature for the concrete distribution; time-dependent $\pi_{ij}(t_{ij})$ denotes the probability of retaining edge $(i,j,t')$. A lower value of $\pi_{ij}(t_{ij})$ indicates the edge is more likely to be a noisy relation and may be discarded during the sampling phase. 
Additionally, we utilize a straight-through method~\cite{bengio_estimating_2013} to select edges with sampling probabilities exceeding a threshold $\varphi_0$.

Based on $\pi_{ij}(t_{ij})$ computing, the $\text{EIB}^{(l)}$ is instantiated as:
\begin{equation}
    \widehat{\text{EIB}}^{(l)} = \mathcal{D}_\text{KL}\left(\text{Ber}\left(\pi_{ij}^{(l)}(t_{ij})\right) \| \text{Ber}\left(\pi_0\right)\right),
    \label{eq:eib_hat}
\end{equation}
where $\pi_{ij}^{(l)}(t_{ij})$ is the edge retention probability for edge $(i,j,t_{ij})$  in layer $ l $ and $\pi_0$ is a hyperparameter of the prior Bernoulli distribution. 

\textbf{(2) Learning Distribution of Representation.}
To model the $\text{XIB}^{(l)}$ as defined in \eqref{eq:xib}, we assume that the node representations follow a Gaussian distribution:
\begin{equation}
    p\bigl(Z^{(l)}(t) \mid G(t)\bigr) = \text{Gauss}(\mu^{(l)}, \sigma^{(l)}).
\end{equation}
Consistent with the approach in~\cite{alemi_deep_2019}, we adopt a fixed $ K $-dimensional spherical Gaussian prior: $q\bigl(Z^{(l)}(t)\bigr) = \text{Gauss}(0, I)$. After TGIB filtering, the refined graph $ G_{\text{IB}}^{(l)}(t) $ is input to the TGN's aggregator to learn the distribution parameters of the node representations:
\begin{equation}
    [\mu^{(l)} \oplus \log \sigma^{(l)}] = \text{AGG}^{(l)}_{\text{TGN}}(G_{\text{IB}}^{(l)}(t)),
    \label{eq:tgn_embedding}
\end{equation}
where $\mu^{(l)} \in \mathbb{R}^{K \times 1}$ and $\sigma^{(l)} \in \mathbb{R}^{K \times 1}$ are the mean and standard deviation vectors generated at layer $ l $. The function $\text{AGG}^{(l)}_{\text{TGN}}$ denotes the message passing and aggregation mechanism of the TGN. Alternatively, the CAW model can substitute the TGN:
\begin{equation}
    [\mu \oplus \log \sigma] = \text{AGG}_{\text{CAW}}(G_{\text{IB}}(t)),
\end{equation}
where $\text{AGG}_{\text{CAW}}$ is the random walk and aggregation of CAW.

\begin{table*}[htb]
\centering
\caption{TLP performance (average precision, mean $\pm$ std). Best results are marked in bold, second best results are underlined.}
\label{tab:model_performance}
\begin{tabular}{@{}lccccc}
\toprule
\textbf{Setting} & \textbf{Model} & \textbf{UCI} & \textbf{Social Evolution}& \textbf{MOOC} & \textbf{Wikipedia} \\
\midrule
\multirow{11}{*}{\rotatebox[origin=c]{90}{Transductive}}& JODIE           & 86.73 $\pm$ 1.0 & 76.74 $\pm$ 1.2& 79.98 $\pm$ 0.4& 94.62 $\pm$ 0.5 \\
& DyRep           & 54.60 $\pm$ 3.1 & 66.02 $\pm$ 0.7& 80.45 $\pm$ 0.5& 94.59 $\pm$ 0.2 \\
& TGAT            & 77.51 $\pm$ 0.7 & 63.36 $\pm$ 0.2& 69.75 $\pm$ 0.2& 95.34 $\pm$ 0.1 \\
& GraphMixer &  93.40 $\pm$ 0.6 &  89.11  $\pm$  0.2& 82.73 $\pm$ 0.2& 96.89 $\pm$ 0.1\\
& GraphMixer+TGSL & 88.94 $\pm$ 0.9 & 89.30 $\pm$ 0.2& 79.59 $\pm$ 0.7& 97.19 $\pm$ 0.4 \\
& DyGFormer  & \underline{95.76 $\pm$ 0.2} & 90.75  $\pm$  0.2& 87.23  $\pm$ 0.5 & 98.82 $\pm$ 0.1 \\
& FreeDyG  &   \textbf{96.01 $\pm$ 0.1}& \underline{90.78  $\pm$  0.1} & \underline{89.61 $\pm$ 0.3} &  \textbf{99.22 $\pm$ 0.1}\\
\cmidrule(lr){2-6}
& TGN             & 80.40  $\pm$  1.4 & 87.69 $\pm$ 0.5& 89.26 $\pm$ 0.5 & 98.46 $\pm$ 0.1 \\
& GTGIB-TGN        & 87.04 $\pm$ 0.5 & \textbf{90.82 $\pm$ 0.2}&  89.44 $\pm$ 0.3&  98.55 $\pm$ 0.1                 \\
\cmidrule(lr){2-6}
& CAW             & 92.16 $\pm$ 0.1 & 88.68 $\pm$ 0.5& 83.56  $\pm$  0.4& 98.63 $\pm$ 0.1 \\
& GTGIB-CAW  & 94.03 $\pm$ 0.1& 88.73 $\pm$ 0.2&  \textbf{92.15 $\pm$ 0.3}& \underline{98.95 $\pm$ 0.1} \\
\midrule
\multirow{11}{*}{\rotatebox[origin=c]{90}{Inductive}}& JODIE           & 75.26 $\pm$ 1.7 & 78.53 $\pm$ 1.8& 79.43 $\pm$ 0.7& 93.11 $\pm$ 0.4 \\
& DyRep           & 50.96 $\pm$ 1.9 & 67.65 $\pm$ 1.0& 78.98 $\pm$ 0.4& 92.05 $\pm$ 0.3 \\
& TGAT            & 70.54 $\pm$ 0.5 & 61.37 $\pm$ 0.3& 72.51 $\pm$ 0.2& 93.99 $\pm$ 0.3 \\
& GraphMixer &   91.19 $\pm$ 0.4  &  86.01  $\pm$  0.1&  81.41 $\pm$ 0.2 & 96.12 $\pm$ 0.2\\
& GraphMixer+TGSL &  88.01 $\pm$  0.1 & 86.19 $\pm$ 1.3& 77.15 $\pm$ 0.6& 96.56 $\pm$ 0.4 \\
& DyGFormer  &  94.38 $\pm$ 0.1 & \underline{89.89  $\pm$  0.2}& 86.96  $\pm$  0.5& 98.59 $\pm$ 0.1\\
& FreeDyG  &  \underline{94.81  $\pm$ 0.1} & 88.84  $\pm$  0.1&  87.75 $\pm$  0.6&  \underline{98.85 $\pm$ 0.0}\\
\cmidrule(lr){2-6}
& TGN             & 76.70 $\pm$ 0.9& 85.27 $\pm$ 0.5& 89.01 $\pm$ 0.8& 97.81 $\pm$ 0.1 \\
& GTGIB-TGN        &  81.54 $\pm$ 0.5& 87.26 $\pm$ 0.4&  \underline{89.29 $\pm$ 0.6}&   98.13 $\pm$ 0.1                \\
\cmidrule(lr){2-6}
& CAW             & 93.56 $\pm$ 0.1 & 88.67 $\pm$ 0.5&  80.86 $\pm$ 0.3& 98.52 $\pm$ 0.1\\
& GTGIB-CAW        & \textbf{94.95 $\pm$ 0.1} & \textbf{92.75 $\pm$ 0.3} & \textbf{90.77 $\pm$ 0.2} & \textbf{99.05 $\pm$ 0.1}\\
\bottomrule
\end{tabular}
\end{table*}

\textbf{(3) Sampling Representation.}
To derive the differentiable node representations $Z^{(l)}(t)$ from the distribution, we utilize reparameterization~\cite{kingma_auto-encoding_2013}:
\begin{equation}
    Z^{(l)}(t) = \mu^{(l)} + \sigma^{(l)} \odot \varepsilon,
\end{equation}
where $\varepsilon$ is sampled from a standard Gaussian distribution. 
Thus we instantiate the $\text{XIB}^{(l)}$ as:
\begin{align}
\widehat{\text{XIB}}^{(l)} = \mathcal{D}_\text{KL}\big(&\text{Gauss}(\mu^{(l)}, \sigma^{(l)} ) \| 
\text{Gauss}(0, I)\big).
\label{eq:xib_hat}
\end{align}
\textbf{(4) Overall Training Objective.}
As shown in Algorithm~\ref{alg:tgib}, $q\bigl(Y_i \mid Z^{(L)}_i(t)\bigr)$ in \eqref{eq:upper_bound1} will be instantiated as a trainable distribution,  parameterized by the temporal graph learning model. Therefore, the overall training objective is:
\begin{equation}
    \mathcal{L}_{\text{TGIB}} = \mathcal{L}_{\text{CE}}(\hat{Y}, Y) + \sum_{l=1}^{L} \left( \alpha \widehat{\text{EIB}}^{(l)} + \beta \widehat{\text{XIB}}^{(l)} \right),
    \label{eq:overall loss}
\end{equation}
 where $\mathcal{L}_{\text{CE}}$ is the cross-entropy loss,  $\alpha$ and $\beta$ are hyperparameters of the $\widehat{\text{EIB}}^{(l)}$ and $\widehat{\text{XIB}}^{(l)}$ regularization.

\section{Experiments}\label{sec:experiments}

\textbf{Datasets.} We adopt four real-world datasets, Wikipedia, MOOC, Social Evolution and UCI \cite{rossi_network_2015,kumar_predicting_2019,xu_inductive_2020,wang_inductive_2022}. Section~\ref{appendix:dataset} of the appendix provides a detailed description and statistics of the datasets. We select the 2-week Social Evolution dataset in experiments. All datasets employ zero vectors for node features. 

\textbf{Baseline.} We compare our proposed GTGIB-TGN and GTGIB-CAW with existing state-of-the-art methods. The temporal graph learning baselines include \textbf{JODIE}~\cite{kumar_predicting_2019}, \textbf{DyRep}~\cite{trivedi_dyrep_2019}, \textbf{TGAT}~\cite{xu_inductive_2020}, \textbf{TGN}~\cite{rossi_temporal_2020}, \textbf{CAW}~\cite{wang_inductive_2022}, \textbf{GraphMixer}~\cite{cong_we_2023}, \textbf{DyGFormer}~\cite{yu_towards_2023} and \textbf{FreeDyG}~\cite{tian_freedyg_2024}. Additionally, we include the temporal graph structure learning baseline \textbf{GraphMixer+TGSL}~\cite{zhang_time-aware_2023}. The TGN model employs attention-based aggregation. The DyGFormer and FreeDyG use random negative sampling. We exclude GSL-based algorithms such as CGI~\cite{wei_contrastive_2022} and IB-based approaches like VIB-GSL~\cite{sun_graph_2022} and SIB~\cite{yu_graph_2020} as they are designed for static and graph-level tasks. All results were obtained using implementations and guidelines from the official repositories. Details are in the Appendix, Section~\ref{appendix:baselines}.

\textbf{Implementation Details.} Experiments use a $70$\%-$15$\%-$15$\% (train-val-test) temporal split, with average precision (AP) as the metric. Results are reported for transductive (between seen nodes in training) and inductive (between unseen nodes) predictions, showing the mean and standard deviation of AP over 5 runs. All methods are implemented in PyTorch on an NVIDIA V100 GPU. 
Following prior work~\cite{rossi_temporal_2020} and for simplicity, $ [0,t]$ is used as the observation window. Batch size is 200; negative samples size is 5; $\tau = 0.7$; $\pi_0 = 0.5$; $\varphi_0 = 0.1$; $K$ matches the node feature dimension of each dataset. Hyperparameter search is conducted for $\alpha \in \{10,1,0.1\}$ and $\beta \in \{10^{-3}, 10^{-4}, \dots, 10^{-10}\}$. The random sampling enhancer samples 10 nodes, while the hop-based sampling enhancer samples 20 nodes (10 nodes from the 1-hop and 10 nodes from 2-hop neighbors). TGN and CAW were chosen as backbones for the distinct paradigms (message-passing for TGN; random walk for CAW). All results are averages over 5 runs. In tables showing results, the best results are in \textbf{bold} and the second best are \underline{underlined}.

\subsection{Results}
Table~\ref{tab:model_performance} compares models for TLP in the transductive setting, which predicts future links between previously seen nodes and the inductive setting, which predicts links involving unseen nodes. The evaluation is conducted on the UCI, Social Evolution, MOOC and Wikipedia datasets. From Table~\ref{tab:model_performance}, we observe several key findings:

\textbf{(1)} GTGIB demonstrates remarkable improvements over both TGN and CAW across all four datasets in the inductive setting. Most notably, even with backbones weaker than FreeDyG or DyGFormer, GTGIB-CAW consistently outperforms all existing baseline methods across UCI, Social Evolution, MOOC and Wikipedia datasets. GTGIB achieves average performance gains of 2.32\% and 4.72\% over TGN and CAW, respectively in the inductive setting. Furthermore, we observe consistently reduced variance in the average precision scores for both TGN and CAW variants when enhanced with GTGIB, which shows that GTGIB enhances representation and prediction stability. 

\textbf{(2)} GTGIB demonstrates effectiveness in transductive settings, with GTGIB-TGN outperforming baselines on Social Evolution and GTGIB-CAW excelling on MOOC datasets. Across all datasets, GTGIB improves TGN and CAW by 3.03\% and 3.17\%, respectively. In cases where GTGIB 
models do not achieve state-of-the-art, this reflects the rich information in transductive settings where different architectural (frequency-based FreeDyG, MP-based TGN, RW-based CAW and transformer-based DyGFormer) become dominant factors in determining model performance. 

\textbf{(3)} GTGIB-TGN and GTGIB-CAW's consistent improvements across datasets demonstrate GTGIB's versatility. By refining graph structure rather than model architecture, GTGIB could be integrated with diverse approaches, ranging from message-passing-based TGN to random-walk-based CAW and beyond, showcasing its flexibility and generalisability. 

\subsection{Ablation Study}
\begin{table}[b]
\centering

\caption{Ablation study of GTGIB-TGN on UCI. ``Rand only" means only the random structure enhancer is used; ``hop only" means only the hop-based structure enhancer is used; 
``\textit{w/o} enhancer" means no sampling is used. 
``\textit{w/o} TGIB" means both random and hop-based sampling but no TGIB is used.
``$\Delta$ mean" represents the difference between the mean values of each model and TGN.}
\label{tab:ablation tgib-tgn}
\resizebox{0.5\textwidth}{!}{
\begin{tabular}{lcc|cc}
\toprule
\textbf{Model} & \textbf{Transductive AP} & \textbf{$\Delta$ mean} & \textbf{Inductive AP} & \textbf{$\Delta$ mean} \\
\midrule
TGN & 80.40 $\pm$ 1.4 & -- & 76.70 $\pm$ 0.9 & -- \\
GTGIB-TGN (rand only) & 83.63 $\pm$ 1.1 & +3.23 & 79.29 $\pm$ 1.0 & +2.59 \\
GTGIB-TGN (hop only) & 85.08 $\pm$ 0.8 & +4.68 & \underline{79.80 $\pm$ 0.5} & \underline{+3.10} \\
GTGIB-TGN (\textit{w/o.} enhancer) & 82.94 $\pm$ 0.5 & +2.54 & 74.47 $\pm$ 0.4 & -2.23 \\
GTGIB-TGN (\textit{w/o.} TGIB) & \underline{85.88 $\pm$ 0.6} & \underline{+5.48} & 79.78 $\pm$ 0.5 & +3.08 \\
\cmidrule(lr){1-5}
GTGIB-TGN & \textbf{87.04} $\pm$ 0.5 & \textbf{+6.64} & \textbf{81.54} $\pm$ 0.5 & \textbf{+4.84} \\
\bottomrule
\end{tabular}
}
\end{table}

The ablation study on the UCI dataset demonstrates the contribution of each component in the GTGIB-TGN model (see Table~\ref{tab:ablation tgib-tgn}).

In transductive setting, the random sampling enhancer (GTGIB-TGN (rand only)) and hop-based sampling enhancer (GTGIB-TGN (hop only)) individually improved the Average Precision by 3.23 and 4.68 points, respectively. Using both enhancers (GTGIB-TGN (\textit{w/o} TGIB)) further increased the Average Precision by 5.48 points, indicating their complementary nature. TGIB filtering alone (GTGIB-TGN (\textit{w/o} enhancer)) also improved the AP by 2.54 points. The complete GTGIB-TGN model, combining TGIB filtering with both enhancers, achieved the highest improvement of 6.64 points.

In the inductive setting, similar trends were observed, with the random and hop-based sampling enhancers improving the AP by 2.59 and 3.10 points, respectively. The combination of both enhancers (GTGIB-TGN (\textit{w/o} TGIB)) increased the AP by 3.08 points. The GTGIB-TGN model achieved the most significant improvement of 4.84 points, emphasizing the effectiveness of TGIB filtering and enhancers combined. Interestingly, TGIB filtering alone (GTGIB-TGN (\textit{w/o.} enhancer)) decreased the AP by 2.23 points, suggesting that enhancers play a more crucial role in the inductive setting.

\subsection{Hyperparameter Analysis of EIB and XIB}

\begin{figure}[htbp]
    \centering    
    \subfloat[$\alpha$-Inductive]
    {
        \begin{minipage}[t]{0.23\textwidth}
            \centering    
            \includegraphics[width=1\textwidth]{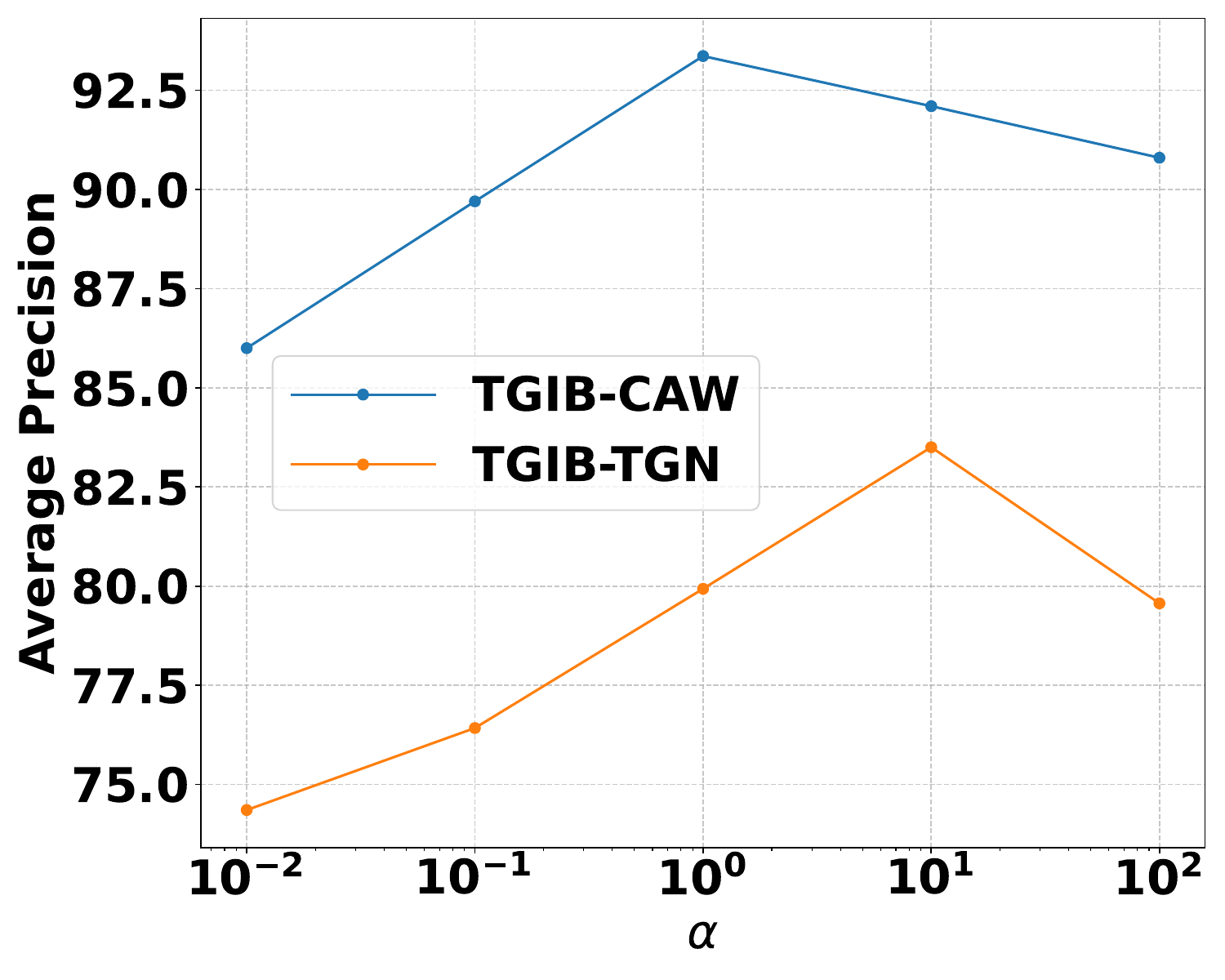}   
        \end{minipage}
    }
    \subfloat[$\alpha$-Transductive] 
    {
        \begin{minipage}[t]{0.23\textwidth}
            \centering    
            \includegraphics[width=1\textwidth]{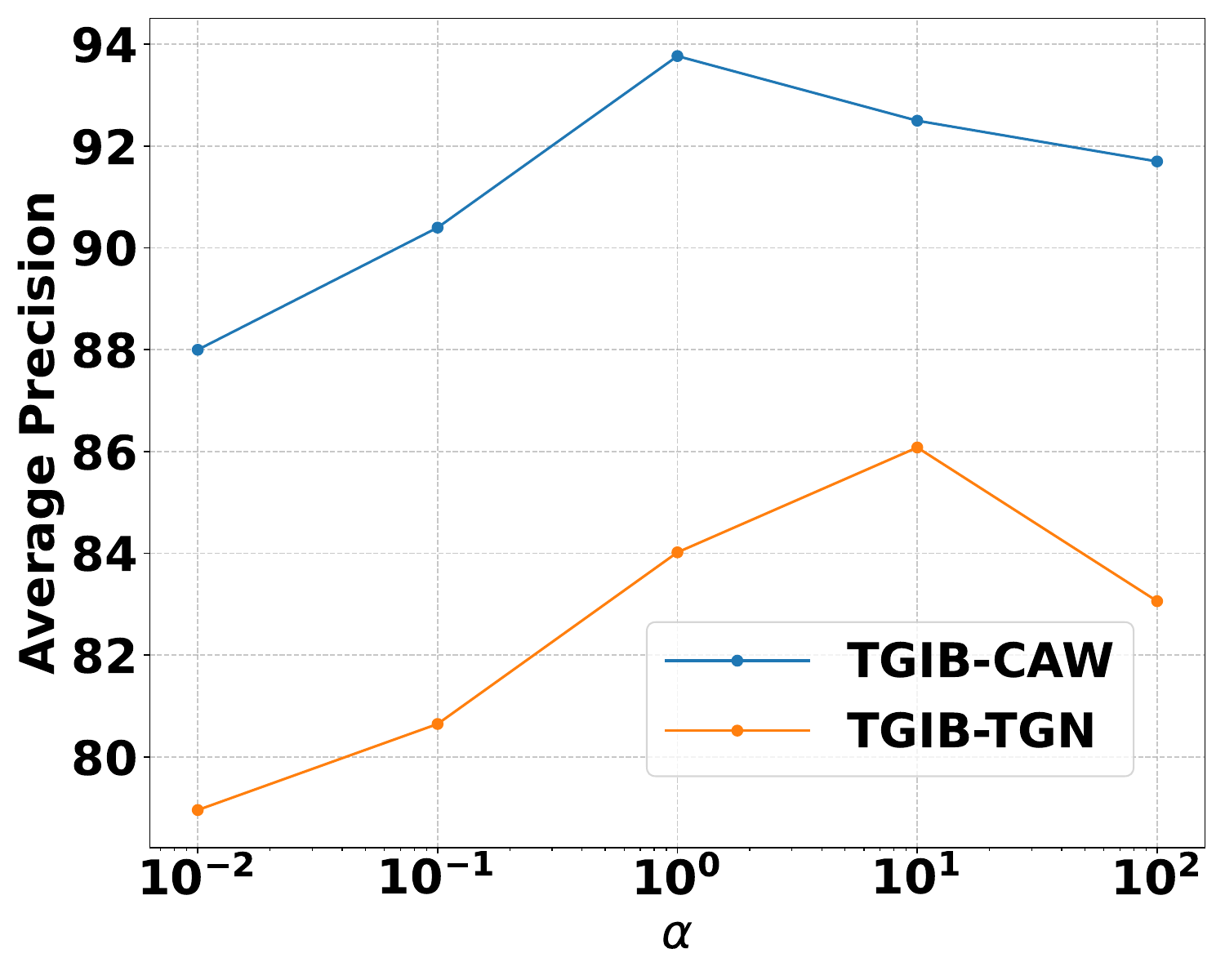}   
        \end{minipage}
    }
    
    \subfloat[$\beta$-Inductive] 
    {
        \begin{minipage}[t]{0.23\textwidth}
            \centering      
            \includegraphics[width=1\textwidth]{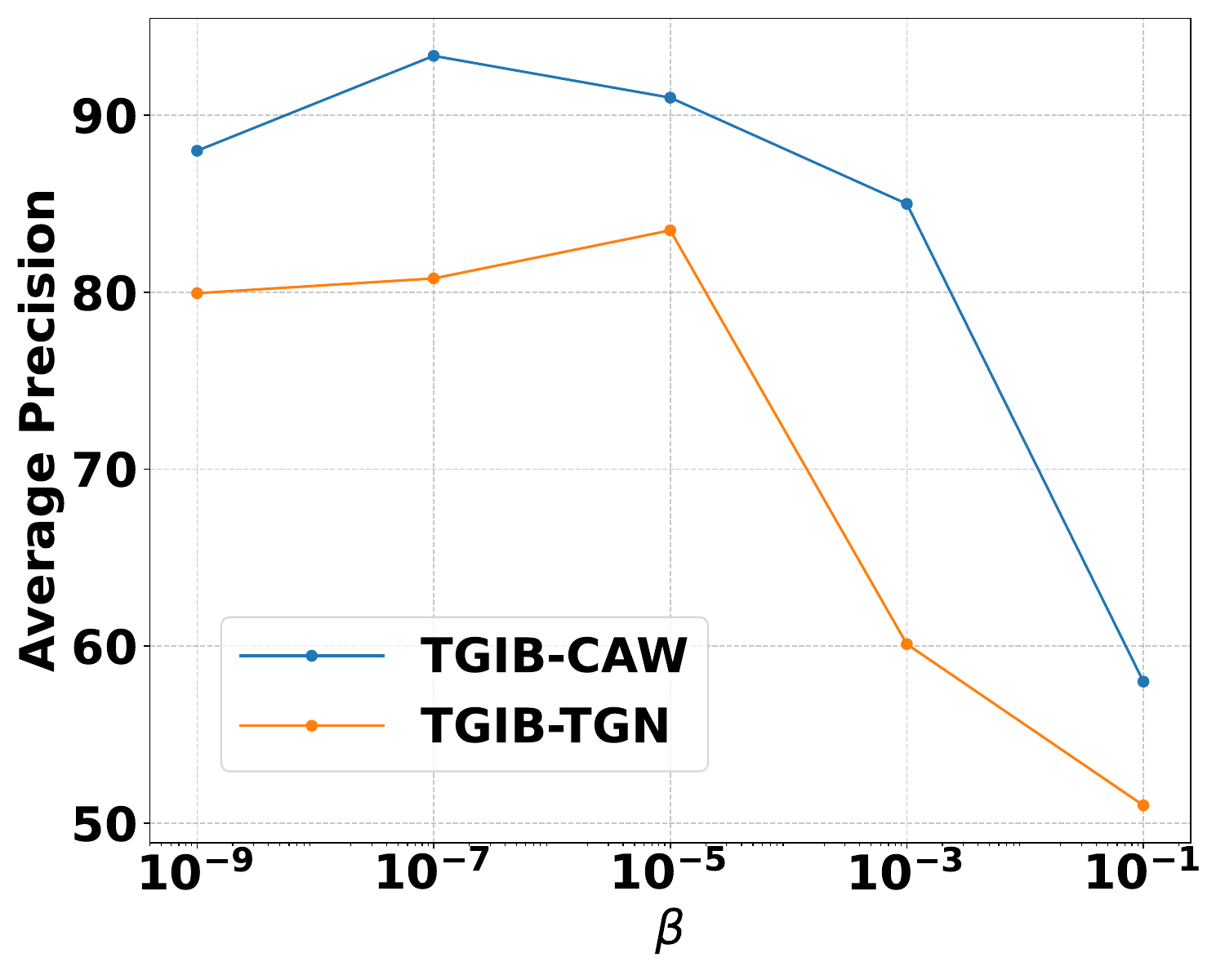}   
        \end{minipage}
    }
    \subfloat[$\beta$-Transductive] 
    {
        \begin{minipage}[t]{0.23\textwidth}
            \centering      
            \includegraphics[width=1\textwidth]{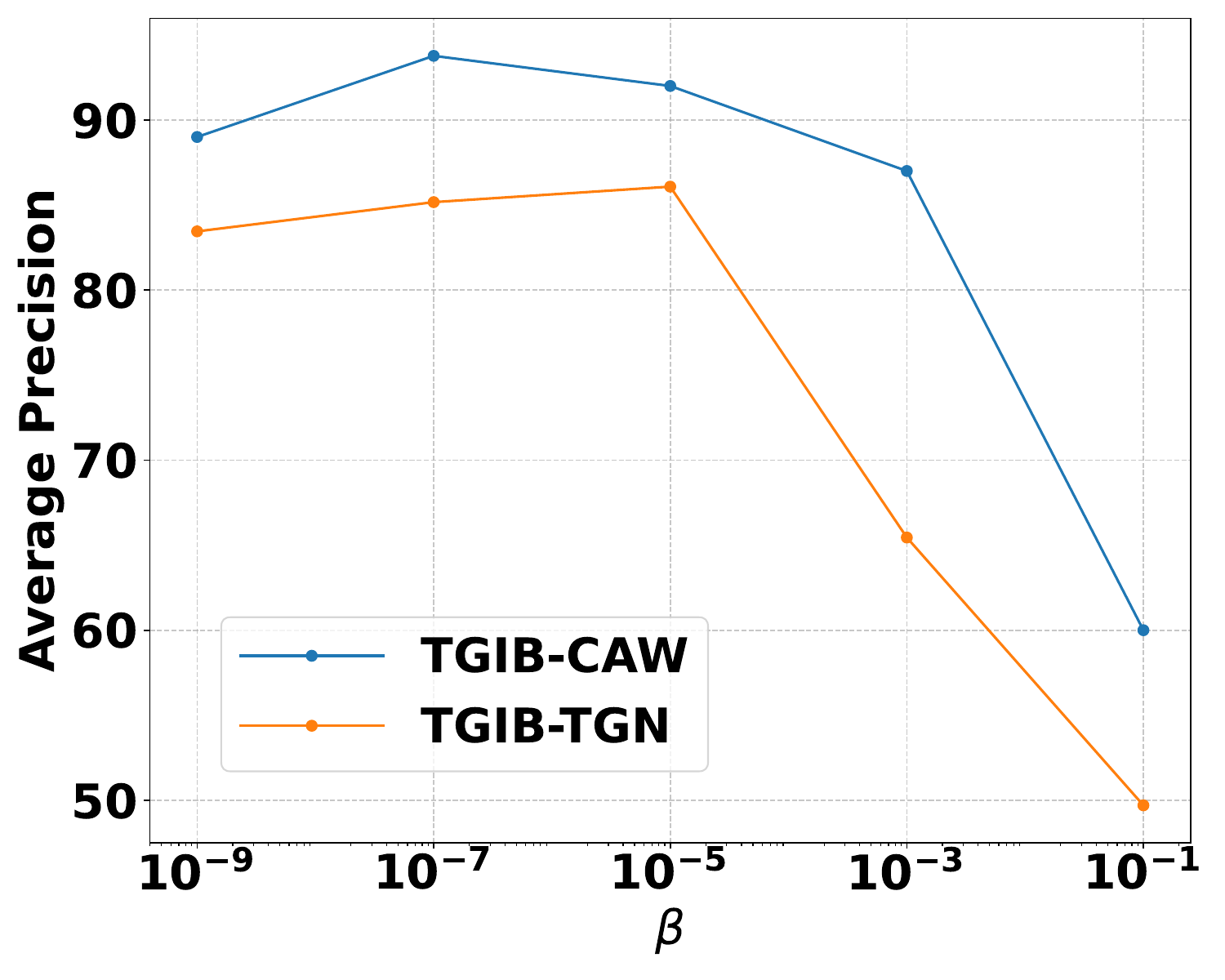}   
        \end{minipage}
    }
    \caption{$\alpha$ and $\beta$ analysis of GTGIB-TGN on the UCI.} 
    \label{fig:param_analysis}  
\end{figure}

We investigate the impact of the Lagrangian multipliers EIB's  $ \alpha $ and XIB's $ \beta $. Using the optimal parameter combinations identified for GTGIB-CAW with $\{\alpha = 1, \beta = 10^{-7}\}$ and GTGIB-TGN with $\{\alpha = 10, \beta = 10^{-5}\}$ on the UCI dataset as baselines, we analyze the effect of varying one parameter while keeping the other fixed. 

\textbf{(1) Differences between $\alpha$ and $\beta$.} EIB focuses on graph structure (edges) while XIB regularizes feature information (node embeddings). Lower hyperparameter values indicate less task-irrelevant information. Figure~\ref{fig:param_analysis} shows the optimal $\alpha$ value of 1 or 10 for TGN, and $\beta$ values exceeding $10^{-3}$ this cause over-regularization and performance degradation. GTGIB uses a larger $\alpha$ for EIB because of abundant structural noise from both the original dataset and our structure enhancer. Conversely, GTGIB uses a much smaller $\beta$ for XIB because the feature information contains much less task-irrelevant content—we adopt zero vectors for initial node features (in line with inductive learning practices), and the node embeddings are learned entirely under the TLP via strong backbones. 

\textbf{(2)} \textbf{Hump-shaped trend.} Across all datasets, AP plotted against $ \alpha $ and $ \beta $ forms a hump-shaped surface, showing that the two parameters peak at different values. Figure~\ref{fig:param_analysis} reveals that the best $ \beta $ values are relatively small compared to $\alpha$; high values of $ \beta$ reduce performance, which suggests that $\beta$ still has a significant effect.  

Additionally, for GTGIB-TGN the optimal pair is $\{\alpha = 10, \beta = 10^{-5}\}$ on UCI, whereas Social Evolution prefers $\{\alpha = 10, \beta = 10^{-8}\}$. The relatively small $\beta$ value reflects differing levels of task-irrelevant information. Social Evolution has a mean degree of 1013.59 (see Appendix, Section~\ref{appendix:dataset}), about 15 times UCI; its nodes therefore possess far richer interaction histories, producing cleaner embeddings with naturally lower noise, so a small $\beta$ value is sufficient.

\subsection{Analysis of Structure Enhancer Sampling Size}\label{sec:sampling_size}
\begin{table}[htbp]
  \centering
  \caption{GTGIB-TGN on UCI. ``\#sample'' is the total number of neighbors sampled by the structure enhancer. $\Delta$ over the last line.}
  \label{tab:sampling_size}
  \begin{tabular}{r r r r r}
    \toprule
    \#samples & Transductive AP & $\Delta$ & Inductive AP & $\Delta$ \\
    \midrule
    0  & 82.94 & –    & 74.47 & –    \\
    15 & 83.74 & 0.80 & 76.11 & 1.64 \\
    30 & 86.04 & 2.30 & 81.51 & 5.40 \\
    45 & 88.96 & 2.92 & 84.56 & 3.05 \\
    60 & 92.29 & 3.33 & 87.36 & 2.81 \\
    75 & 92.64 & 0.35 & 87.67 & 0.31 \\
    \bottomrule
  \end{tabular}
\end{table}

Proofs of structure enhancer in the Appendix, Section~\ref{appendix:structure enhancer} state that once the sample size passes a given threshold, the ground-truth edge has a non-zero sampling probability. Suppose the probability is fixed; it is akin to a Bernoulli process. In Table~\ref{tab:sampling_size}, despite the UCI dataset comprising 1899 nodes, AP gains jump after 30 samples and level off at 75, implying the UCI threshold for GTGIB-TGN lies below 30. The trend of observed $\Delta$ improvements is also consistent with the Bernoulli assumption, supporting our theory. It shows that the more we sample, the better the prediction performance we get, consistently outperforming no-sampling settings.

\subsection{Comparison with Other IB-based Methods}

\begin{table}[htbp]
\caption{Comparison of IB-based methods in TLP (AUC, mean $\pm$ std)}
\centering
\resizebox{0.5\textwidth}{!}{
\begin{tabular}{l|c|c}
\hline
\textbf{Model} & \textbf{Transductive AUC (Yelp)} & \textbf{Inductive AUC (Yelp)} \\
\hline
GIB           & 77.52 $\pm$ 0.4& – \\
DGIB-Bern     & 76.88 $\pm$ 0.2& – \\
DGIB-Cat      & 79.53 $\pm$ 0.2& – \\
\hline
GTGIB-TGN     & \textbf{90.23 $\pm$ 0.1}& \textbf{82.94 $\pm$0.2} \\
GTGIB-CAW     & 88.51 $\pm$ 0.1& 73.83 $\pm$0.3\\
\hline
\end{tabular}
}

\label{tab:ib_results}
\end{table}

Node‐level IB methods like GIB~\cite{wu_graph_2020} and DGIB~\cite{yuan_dynamic_2024} operate on DTDG and therefore cannot be evaluated on our CTDG datasets. However, only the DTDG‐processed versions of datasets are publicly released from DGIB and cannot be converted to CTDG. To enable a fair comparison, we obtained the same source dataset as DGIB, Yelp~\cite{sankar_dysat_2020}, and transformed it into CTDG to evaluate our models based on the same AUC metric. As Table~\ref{tab:ib_results} shows, GTGIB-TGN and GTGIB-CAW achieve significantly higher AUC than GIB and DGIB, and additionally support inductive tasks.

\section{Conclusion}\label{sec:conclusion}

The paper presents GTGIB, a versatile framework integrating graph structure learning with temporal graph information bottleneck principles to advance inductive representation learning in temporal networks. Experiments demonstrate GTGIB's effectiveness and versatility across TGN and CAW on inductive and transductive learning. Future work could explore GTGIB's robustness and integration with other temporal graph learning architectures. We believe GTGIB will inspire further innovations in temporal graph representation learning.



\begin{thebibliography}{45}
\providecommand{\natexlab}[1]{#1}
\providecommand{\url}[1]{\texttt{#1}}
\expandafter\ifx\csname urlstyle\endcsname\relax
  \providecommand{\doi}[1]{doi: #1}\else
  \providecommand{\doi}{doi: \begingroup \urlstyle{rm}\Url}\fi

\bibitem[Adjeisah et~al.(2023)Adjeisah, Zhu, Xu, and Ayall]{adjeisah_towards_2023}
M.~Adjeisah, X.~Zhu, H.~Xu, and T.~A. Ayall.
\newblock Towards data augmentation in graph neural network: {An} overview and evaluation.
\newblock \emph{Computer Science Review}, 47:\penalty0 100527, 2023.

\bibitem[Alemi et~al.(2019)Alemi, Fischer, Dillon, and Murphy]{alemi_deep_2019}
A.~A. Alemi, I.~Fischer, J.~V. Dillon, and K.~Murphy.
\newblock Deep {Variational} {Information} {Bottleneck}, Oct. 2019.
\newblock URL \url{10.48550/arXiv.1612.00410}.

\bibitem[Anderson(1991)]{anderson_continuous-time_1991}
W.~J. Anderson.
\newblock \emph{Continuous-{Time} {Markov} {Chains}}.
\newblock Springer {Series} in {Statistics}. Springer New York, New York, NY, 1991.

\bibitem[Battaglia et~al.(2018)Battaglia, Hamrick, Bapst, Sanchez-Gonzalez, Zambaldi, Malinowski, Tacchetti, Raposo, Santoro, Faulkner, Gulcehre, Song, Ballard, Gilmer, Dahl, Vaswani, Allen, Nash, Langston, Dyer, Heess, Wierstra, Kohli, Botvinick, Vinyals, Li, and Pascanu]{battaglia_relational_2018}
P.~W. Battaglia, J.~B. Hamrick, V.~Bapst, A.~Sanchez-Gonzalez, V.~Zambaldi, M.~Malinowski, A.~Tacchetti, D.~Raposo, A.~Santoro, R.~Faulkner, C.~Gulcehre, F.~Song, A.~Ballard, J.~Gilmer, G.~Dahl, A.~Vaswani, K.~Allen, C.~Nash, V.~Langston, C.~Dyer, N.~Heess, D.~Wierstra, P.~Kohli, M.~Botvinick, O.~Vinyals, Y.~Li, and R.~Pascanu.
\newblock Relational inductive biases, deep learning, and graph networks, 2018.
\newblock URL \url{https://arxiv.org/abs/1806.01261}.

\bibitem[Bengio et~al.(2013)Bengio, Léonard, and Courville]{bengio_estimating_2013}
Y.~Bengio, N.~Léonard, and A.~Courville.
\newblock Estimating or propagating gradients through stochastic neurons for conditional computation, 2013.
\newblock URL \url{https://arxiv.org/abs/1308.3432}.

\bibitem[Chen and Ying(2024)]{chen_tempme_2024}
J.~Chen and R.~Ying.
\newblock Tempme: {Towards} the explainability of temporal graph neural networks via motif discovery.
\newblock \emph{Advances in Neural Information Processing Systems}, 36, 2024.

\bibitem[Chen et~al.(2021)Chen, Zhang, Xu, Fu, Zhang, Zhang, and Xuan]{chen_e-lstm-d_2021}
J.~Chen, J.~Zhang, X.~Xu, C.~Fu, D.~Zhang, Q.~Zhang, and Q.~Xuan.
\newblock E-{LSTM}-{D}: {A} {Deep} {Learning} {Framework} for {Dynamic} {Network} {Link} {Prediction}.
\newblock \emph{IEEE Transactions on Systems, Man, and Cybernetics: Systems}, 51\penalty0 (6):\penalty0 3699--3712, June 2021.

\bibitem[Cong et~al.(2023)Cong, Zhang, Kang, Yuan, Wu, Zhou, Tong, and Mahdavi]{cong_we_2023}
W.~Cong, S.~Zhang, J.~Kang, B.~Yuan, H.~Wu, X.~Zhou, H.~Tong, and M.~Mahdavi.
\newblock Do we really need complicated model architectures for temporal networks?, 2023.
\newblock URL \url{https://arxiv.org/abs/2302.11636}.

\bibitem[Goyal et~al.(2018)Goyal, Kamra, He, and Liu]{goyal_dyngem_2018}
P.~Goyal, N.~Kamra, X.~He, and Y.~Liu.
\newblock Dyngem: Deep embedding method for dynamic graphs, 2018.
\newblock URL \url{https://arxiv.org/abs/1805.11273}.

\bibitem[Gravina et~al.(2024)Gravina, Lovisotto, Gallicchio, Bacciu, and Grohnfeldt]{gravina_long_2024}
A.~Gravina, G.~Lovisotto, C.~Gallicchio, D.~Bacciu, and C.~Grohnfeldt.
\newblock Long {Range} {Propagation} on {Continuous}-{Time} {Dynamic} {Graphs}.
\newblock In \emph{Proceedings of the 41st {International} {Conference} on {Machine} {Learning}}, volume 235, pages 16206--16225. PMLR, July 2024.

\bibitem[Higgins et~al.(2017)Higgins, Matthey, Pal, Burgess, Glorot, Botvinick, Mohamed, and Lerchner]{higgins_beta-vae_2017}
I.~Higgins, L.~Matthey, A.~Pal, C.~Burgess, X.~Glorot, M.~Botvinick, S.~Mohamed, and A.~Lerchner.
\newblock beta-{VAE}: Learning basic visual concepts with a constrained variational framework.
\newblock In \emph{International Conference on Learning Representations}, 2017.

\bibitem[Holme and Saramäki(2012)]{holme_temporal_2012}
P.~Holme and J.~Saramäki.
\newblock Temporal networks.
\newblock \emph{Physics Reports}, 519\penalty0 (3):\penalty0 97--125, Oct. 2012.

\bibitem[Jang et~al.(2017)Jang, Gu, and Poole]{jang_categorical_2017}
E.~Jang, S.~Gu, and B.~Poole.
\newblock Categorical reparameterization with gumbel-softmax, 2017.
\newblock URL \url{https://arxiv.org/abs/1611.01144}.

\bibitem[Jiang(2022)]{jiang_graph-based_2022}
W.~Jiang.
\newblock Graph-based deep learning for communication networks: {A} survey.
\newblock \emph{Computer Communications}, 185:\penalty0 40--54, Mar. 2022.

\bibitem[Jin et~al.(2020)Jin, Ma, Liu, Tang, Wang, and Tang]{jin_graph_2020}
W.~Jin, Y.~Ma, X.~Liu, X.~Tang, S.~Wang, and J.~Tang.
\newblock Graph {Structure} {Learning} for {Robust} {Graph} {Neural} {Networks}.
\newblock In \emph{Proceedings of the 26th {ACM} {SIGKDD}}, {KDD}'20, pages 66--74, New York, NY, USA, 2020. Association for Computing Machinery.

\bibitem[Kingma and Welling(2022)]{kingma_auto-encoding_2013}
D.~P. Kingma and M.~Welling.
\newblock Auto-encoding variational bayes, 2022.
\newblock URL \url{https://arxiv.org/abs/1312.6114}.

\bibitem[Kipf and Welling(2017)]{kipf_semi-supervised_2017}
T.~N. Kipf and M.~Welling.
\newblock Semi-supervised classification with graph convolutional networks, 2017.
\newblock URL \url{https://arxiv.org/abs/1609.02907}.

\bibitem[Kovanen et~al.(2013)Kovanen, Kaski, Kertész, and Saramäki]{kovanen_temporal_2013}
L.~Kovanen, K.~Kaski, J.~Kertész, and J.~Saramäki.
\newblock Temporal motifs reveal homophily, gender-specific patterns, and group talk in call sequences.
\newblock \emph{Proceedings of the National Academy of Sciences}, 110\penalty0 (45):\penalty0 18070--18075, Nov. 2013.

\bibitem[Kumar et~al.(2019)Kumar, Zhang, and Leskovec]{kumar_predicting_2019}
S.~Kumar, X.~Zhang, and J.~Leskovec.
\newblock Predicting {Dynamic} {Embedding} {Trajectory} in {Temporal} {Interaction} {Networks}.
\newblock In \emph{Proceedings of the 25th {ACM} {SIGKDD}}, {KDD}'19, pages 1269--1278, New York, NY, USA, July 2019. Association for Computing Machinery.

\bibitem[Liu et~al.(2022)Liu, Zheng, Zhang, Chen, Peng, and Pan]{liu_towards_2022}
Y.~Liu, Y.~Zheng, D.~Zhang, H.~Chen, H.~Peng, and S.~Pan.
\newblock Towards {Unsupervised} {Deep} {Graph} {Structure} {Learning}.
\newblock In \emph{Proceedings of the {ACM} {Web} {Conference} 2022}, {WWW}'22, pages 1392--1403, New York, NY, USA, 2022. Association for Computing Machinery.

\bibitem[Marisca et~al.(2024)Marisca, Alippi, and Bianchi]{marisca_graph-based_2024}
I.~Marisca, C.~Alippi, and F.~M. Bianchi.
\newblock Graph-based {Forecasting} with {Missing} {Data} through {Spatiotemporal} {Downsampling}.
\newblock In \emph{Proceedings of the 41st {International} {Conference} on {Machine} {Learning}}, volume 235, pages 34846--34865. PMLR, 2024.

\bibitem[Nguyen et~al.(2018)Nguyen, Lee, Rossi, Ahmed, Koh, and Kim]{nguyen_continuous-time_2018}
G.~H. Nguyen, J.~B. Lee, R.~A. Rossi, N.~K. Ahmed, E.~Koh, and S.~Kim.
\newblock Continuous-{Time} {Dynamic} {Network} {Embeddings}.
\newblock In \emph{Companion {Proceedings} of the {The} {Web} {Conference} 2018}, pages 969--976, 2018.

\bibitem[Rossi et~al.(2020)Rossi, Chamberlain, Frasca, Eynard, Monti, and Bronstein]{rossi_temporal_2020}
E.~Rossi, B.~Chamberlain, F.~Frasca, D.~Eynard, F.~Monti, and M.~Bronstein.
\newblock Temporal {Graph} {Networks} for {Deep} {Learning} on {Dynamic} {Graphs}, Oct. 2020.
\newblock URL \url{http://arxiv.org/abs/2006.10637}.

\bibitem[Rossi and Ahmed(2015)]{rossi_network_2015}
R.~Rossi and N.~Ahmed.
\newblock The {Network} {Data} {Repository} with {Interactive} {Graph} {Analytics} and {Visualization}.
\newblock \emph{Proceedings of the AAAI Conference on Artificial Intelligence}, 29\penalty0 (1), Mar. 2015.

\bibitem[Sankar et~al.(2020)Sankar, Wu, Gou, Zhang, and Yang]{sankar_dysat_2020}
A.~Sankar, Y.~Wu, L.~Gou, W.~Zhang, and H.~Yang.
\newblock {DySAT}: {Deep} {Neural} {Representation} {Learning} on {Dynamic} {Graphs} via {Self}-{Attention} {Networks}.
\newblock In \emph{Proceedings of the 13th {International} {Conference} on {Web} {Search} and {Data} {Mining}}, {WSDM}'20, pages 519--527, New York, NY, USA, 2020. Association for Computing Machinery.

\bibitem[Skarding et~al.(2021)Skarding, Gabrys, and Musial]{skarding_foundations_2021}
J.~Skarding, B.~Gabrys, and K.~Musial.
\newblock Foundations and modeling of dynamic networks using dynamic graph neural networks: {A} survey.
\newblock \emph{IEEE Access}, 9:\penalty0 79143--79168, 2021.

\bibitem[Souza et~al.(2022)Souza, Mesquita, Kaski, and Garg]{souza_provably_2022}
A.~H. Souza, D.~Mesquita, S.~Kaski, and V.~Garg.
\newblock Provably expressive temporal graph networks, Sept. 2022.
\newblock URL \url{https://arxiv.org/abs/2209.15059}.

\bibitem[Sun et~al.(2022)Sun, Li, Peng, Wu, Fu, Ji, and Yu]{sun_graph_2022}
Q.~Sun, J.~Li, H.~Peng, J.~Wu, X.~Fu, C.~Ji, and P.~S. Yu.
\newblock Graph {Structure} {Learning} with {Variational} {Information} {Bottleneck}.
\newblock \emph{Proceedings of the AAAI Conference on Artificial Intelligence}, 36\penalty0 (4):\penalty0 4165--4174, June 2022.

\bibitem[Tian et~al.(2024)Tian, Qi, and Guo]{tian_freedyg_2024}
Y.~Tian, Y.~Qi, and F.~Guo.
\newblock Freedyg: Frequency enhanced continuous-time dynamic graph model for link prediction.
\newblock In \emph{The Twelfth International Conference on Learning Representations}, 2024.
\newblock URL \url{https://openreview.net/forum?id=82Mc5ilInM}.

\bibitem[Tishby et~al.(2000)Tishby, Pereira, and Bialek]{tishby_information_2000}
N.~Tishby, F.~C. Pereira, and W.~Bialek.
\newblock The information bottleneck method, 2000.
\newblock URL \url{https://arxiv.org/abs/physics/0004057}.

\bibitem[Trivedi et~al.(2019)Trivedi, Farajtabar, Biswal, and Zha]{trivedi_dyrep_2019}
R.~Trivedi, M.~Farajtabar, P.~Biswal, and H.~Zha.
\newblock Dyrep: Learning representations over dynamic graphs.
\newblock In \emph{International Conference on Learning Representations}, 2019.
\newblock URL \url{https://openreview.net/forum?id=HyePrhR5KX}.

\bibitem[Vaswani et~al.(2017)Vaswani, Shazeer, Parmar, Uszkoreit, Jones, Gomez, Kaiser, and Polosukhin]{vaswani_attention_2017}
A.~Vaswani, N.~Shazeer, N.~Parmar, J.~Uszkoreit, L.~Jones, A.~N. Gomez, L.~Kaiser, and I.~Polosukhin.
\newblock Attention is all you need.
\newblock \emph{Advances in neural information processing systems}, 30:\penalty0 5998--6008, 2017.

\bibitem[Veličković et~al.(2018)Veličković, Cucurull, Casanova, Romero, Liò, and Bengio]{velickovic_graph_2018}
P.~Veličković, G.~Cucurull, A.~Casanova, A.~Romero, P.~Liò, and Y.~Bengio.
\newblock Graph {Attention} {Networks}, Feb. 2018.
\newblock URL \url{https://arxiv.org/abs/1710.10903}.

\bibitem[Wang et~al.(2022)Wang, Chang, Liu, Leskovec, and Li]{wang_inductive_2022}
Y.~Wang, Y.-Y. Chang, Y.~Liu, J.~Leskovec, and P.~Li.
\newblock Inductive representation learning in temporal networks via causal anonymous walks, 2022.
\newblock URL \url{https://arxiv.org/abs/2101.05974}.

\bibitem[Wei et~al.(2022)Wei, Liang, Liu, and Wang]{wei_contrastive_2022}
C.~Wei, J.~Liang, D.~Liu, and F.~Wang.
\newblock Contrastive {Graph} {Structure} {Learning} via {Information} {Bottleneck} for {Recommendation}.
\newblock In S.~Koyejo, S.~Mohamed, A.~Agarwal, D.~Belgrave, K.~Cho, and A.~Oh, editors, \emph{Advances in {Neural} {Information} {Processing} {Systems}}, volume~35, pages 20407--20420. Curran Associates, Inc., 2022.

\bibitem[Wu et~al.(2022)Wu, Sun, Zhang, Xie, and Cui]{wu_graph_2022}
S.~Wu, F.~Sun, W.~Zhang, X.~Xie, and B.~Cui.
\newblock Graph {Neural} {Networks} in {Recommender} {Systems}: {A} {Survey}.
\newblock \emph{ACM Computing Surveys}, 55\penalty0 (5):\penalty0 97:1--97:37, Dec. 2022.

\bibitem[Wu et~al.(2020)Wu, Ren, Li, and Leskovec]{wu_graph_2020}
T.~Wu, H.~Ren, P.~Li, and J.~Leskovec.
\newblock Graph {Information} {Bottleneck}.
\newblock In H.~Larochelle, M.~Ranzato, R.~Hadsell, M.~F. Balcan, and H.~Lin, editors, \emph{NeurIPS}, volume~33, pages 20437--20448. Curran Associates, Inc., 2020.

\bibitem[Xiong et~al.(2025)Xiong, Zareie, and Sakellariou]{xiong_survey_2025}
J.~Xiong, A.~Zareie, and R.~Sakellariou.
\newblock A survey of link prediction in temporal networks, 2025.
\newblock URL \url{https://arxiv.org/abs/2502.21185}.

\bibitem[Xu et~al.(2020)Xu, Ruan, Korpeoglu, Kumar, and Achan]{xu_inductive_2020}
D.~Xu, C.~Ruan, E.~Korpeoglu, S.~Kumar, and K.~Achan.
\newblock Inductive representation learning on temporal graphs, 2020.
\newblock URL \url{https://arxiv.org/abs/2002.07962}.

\bibitem[Yang et~al.(2021)Yang, Wu, Zheng, Niu, Gu, Wang, Cao, and Guo]{yang_heterogeneous_2021}
L.~Yang, F.~Wu, Z.~Zheng, B.~Niu, J.~Gu, C.~Wang, X.~Cao, and Y.~Guo.
\newblock Heterogeneous {Graph} {Information} {Bottleneck}.
\newblock In \emph{Proceedings of the {Thirtieth} IJCAI}, pages 1638--1645, Montreal, Canada, Aug. 2021. International Joint Conferences on Artificial Intelligence Organization.

\bibitem[Yu et~al.(2020{\natexlab{a}})Yu, Zhang, Jiang, Wu, and Yang]{yu_graph-revised_2020}
D.~Yu, R.~Zhang, Z.~Jiang, Y.~Wu, and Y.~Yang.
\newblock Graph-{Revised} {Convolutional} {Network}.
\newblock In \emph{{ECML} {PKDD} 2020, {Ghent}, {Belgium}, {September} 14–18, 2020, {Proceedings}, {Part} {III}}, pages 378--393, Berlin, Heidelberg, 2020{\natexlab{a}}. Springer.

\bibitem[Yu et~al.(2020{\natexlab{b}})Yu, Xu, Rong, Bian, Huang, and He]{yu_graph_2020}
J.~Yu, T.~Xu, Y.~Rong, Y.~Bian, J.~Huang, and R.~He.
\newblock Graph information bottleneck for subgraph recognition, 2020{\natexlab{b}}.
\newblock URL \url{https://arxiv.org/abs/2010.05563}.

\bibitem[Yu et~al.(2023)Yu, Sun, Du, and Lv]{yu_towards_2023}
L.~Yu, L.~Sun, B.~Du, and W.~Lv.
\newblock Towards better dynamic graph learning: New architecture and unified library.
\newblock In \emph{Thirty-seventh Conference on Neural Information Processing Systems}, 2023.
\newblock URL \url{https://openreview.net/forum?id=xHNzWHbklj}.

\bibitem[Yuan et~al.(2024)Yuan, Sun, Fu, Ji, and Li]{yuan_dynamic_2024}
H.~Yuan, Q.~Sun, X.~Fu, C.~Ji, and J.~Li.
\newblock Dynamic graph information bottleneck.
\newblock In \emph{Proceedings of the ACM Web Conference 2024}, WWW'24, page 469–480, New York, NY, USA, 2024. Association for Computing Machinery.

\bibitem[Zhang et~al.(2023)Zhang, Han, Xiao, and Bai]{zhang_time-aware_2023}
H.~Zhang, X.~Han, X.~Xiao, and J.~Bai.
\newblock Time-aware {Graph} {Structure} {Learning} via {Sequence} {Prediction} on {Temporal} {Graphs}.
\newblock In \emph{Proceedings of the 32nd ACM International Conference on Information and Knowledge Management}, CIKM'23, pages 3288--3297, New York, NY, USA, 2023. Association for Computing Machinery.

\end{thebibliography}

\appendix
\onecolumn
{\Huge\noindent\bf Appendix}

\section{Temporal Graph Information Bottleneck}

\subsection{Proof of Lemma~\ref{lemma:nuisance}}\label{appendix:lemma}
We restate \textbf{Lemma 1}: Suppose a temporal graph \(G(t)\) with label \(Y\) decomposes as  
\(G(t)=f(S(G(t)),G_n)\), where \(S(G(t))\) is sufficient for \(Y\) (i.e.\  
\(P(Y\mid G(t))=P(Y\mid S(G(t)))\)) and \(I(G_n;Y)=0\), so \(G_n\) is the  
task-irrelevant nuisance. Writing \(G_{\mathrm{IB}}(t)\) for the IB-refined  
graph gives
\begin{equation}
I\bigl(G_{\mathrm{IB}}(t);G_n\bigr)\le I\bigl(G_{\mathrm{IB}}(t);G(t)\bigr)-I\bigl(G_{\mathrm{IB}}(t);Y\bigr).
\end{equation}
\begin{proof}
We prove the lemma following a similar strategy to~\cite {sun_graph_2022}. 
Suppose $G(t)$ is defined by $Y$ and $G_n$ and $G_{\text{IB}}(t)$ depends on $G_n$ only through $G(t)$. Thus, we can define the Markov Chain:
\begin{equation}
    \langle (Y, G_n) \rightarrow G(t) \rightarrow G_{\text{IB}}(t)\rangle.
    \label{eq:lemma markov}
\end{equation}
\noindent
According to the data processing inequality (DPI), we have:
\begin{align}
I\left(G_{\text{IB}}(t); G(t)\right) 
&\ge 
I\left(G_{\text{IB}}(t); Y,  G_n\right) \nonumber \\
&= I\left(G_{\text{IB}}(t);  G_n\right) 
+
I\left(G_{\text{IB}}(t); Y \mid  G_n\right) \nonumber \\
&= I\left(G_{\text{IB}}(t);  G_n\right) 
+
H\left(Y \mid  G_n\right)
-
H\left(Y \mid  G_n, G_{\text{IB}}(t)\right).
\label{eq:temporal_dpi_new}
\end{align}

Since $G_n$ is a task-irrelevant nuisance and is therefore independent of $Y$, we have $H\bigl(Y \mid  G_n\bigr) = H(Y)$ and $H\bigl(Y \mid  G_n, G_{\text{IB}}(t)\bigr) \leq H\bigl(Y \mid G_{\text{IB}}(t)\bigr)$. Substituting these into \eqref{eq:temporal_dpi_new}, we get:

\begin{align}
I\bigl(G_{\text{IB}}(t); G(t)\bigr)
&\ge
I\bigl(G_{\text{IB}}(t); G_n\bigr)
+
H\bigl(Y \mid  G_n\bigr)
-
H\bigl(Y \mid  G_n; G_{\text{IB}}(t)\bigr)
\nonumber \\
&\ge
I\bigl(G_{\text{IB}}(t);  G_n\bigr)
+
H(Y)
-
H\bigl(Y \mid G_{\text{IB}}(t)\bigr)
\nonumber \\
&=
I\bigl(G_{\text{IB}}(t);  G_n\bigr)
+
I\bigl(G_{\text{IB}}(t); Y\bigr).
\label{eq:temporal_inequality_new}
\end{align}
\noindent
Thus, rearranging \eqref{eq:temporal_inequality_new} yields:
\begin{equation}
I\bigl(G_{\text{IB}}(t);  G_n\bigr)
\le
I\bigl(G_{\text{IB}}(t); G(t)\bigr)
-
I\bigl(G_{\text{IB}}(t); Y\bigr).    
\end{equation}
\end{proof}
\subsection{Proof of Proposition 2}\label{appendix:upper bounds}

For completeness, we restate the proposition on the upper bounds of TGIB as follows:
\begin{equation}
    \text{TGIB} \triangleq \bigl[ -I( Z^{(L)}(t);Y) + \gamma I(Z^{(L)}(t);G([t-\Delta t,t])) \bigr].
\label{eq:tgib appendix}
\end{equation}
For the first term in \eqref{eq:tgib appendix}, we have:
\begin{align}
&-I\bigl(Y;Z^{(L)}(t)\bigr)
\le 
-\frac{1}{N} \sum_{i=1}^{N} \log q\bigl(Y_i \mid Z^{(L)}_i(t)\bigr),
\label{eq:upper_bound1_restate}
\end{align}
For the second term in \eqref{eq:tgib appendix}, we have:
\begin{align}
&I\bigl(Z^{(L)}(t);G([t-\Delta t,t])\bigr) 
\le
\sum_{l=1}^{L} (\mathrm{EIB}^{(l)}+\mathrm{XIB}^{(l)}),
\label{eq:upper_bound2_restate}
\end{align}
where $\mathrm{XIB}^{(l)}$ and $\mathrm{EIB}^{(l)}$ are given by:
\begin{align}
\mathrm{XIB}^{(l)} 
= \mathbb{E} \Bigl[\log \frac{p\bigl(Z^{(l)}(t) \mid G([t-\Delta t,t])\bigr)}{q\bigl(Z^{(l)}(t)\bigr)}\Bigr],
\label{eq:xib_restate}\\
\mathrm{EIB}^{(l)} 
= \mathbb{E} \Bigl[\log \frac{p\bigl(E^{(l)}(t) \mid G([t-\Delta t,t])\bigr)}{q\bigl(E^{(l)}(t)\bigr)}\Bigr],
\label{eq:eib_restate}
\end{align}
with $Y_i$ denoting the corresponding label, $Z^{(L)}_i(t)$ representing the node embedding at layer $L$ based on the temporal neighborhood and $q\bigl(Y_i \mid Z^{(L)}_i(t)\bigr)$ as the variational approximation of the true posterior $p\bigl(Y_i \mid Z^{(L)}_i(t)\bigr)$.
As instantiated in Section~\ref{sec:instantiating}, the distribution of each layer is computed from the graph and the outputs of the preceding layers. Writing $S^{(l)}(t)=\{E^{(k)}(t),Z^{(k)}(t)\}_{1\le k\le l}$, with $S^{(0)}(t)=\emptyset$, the conditional distributions in \eqref{eq:eib_restate} and \eqref{eq:xib_restate} are $p\bigl(E^{(l)}(t)\mid G([t-\Delta t,t]),S^{(l-1)}(t)\bigr)$ and $p\bigl(Z^{(l)}(t)\mid G([t-\Delta t,t]),S^{(l-1)}(t),E^{(l)}(t)\bigr)$, and the expectations are taken over the graph and all layers.

\begin{proof}
Firstly, we prove \eqref{eq:upper_bound1_restate}. Based on the properties of mutual information:
\begin{align}
I\bigl(Y;Z^{(L)}(t)\bigr)
=
\mathbb{E}_{p(Y,Z^{(L)}(t))}\Bigl[\log p\bigl(Y \mid Z^{(L)}(t)\bigr)\Bigr]
-
\mathbb{E}_{p(Y)}\bigl[\log p(Y)\bigr].
\label{eq:MI}
\end{align}
We introduce a variational distribution $q\bigl(Y \mid Z^{(L)}(t)\bigr)$. By the non-negativity of the KL divergence,
\begin{align}
\mathbb{E}\bigl[\log p\bigl(Y \mid Z^{(L)}(t)\bigr)\bigr]
&=
\mathbb{E}\bigl[\log q\bigl(Y \mid Z^{(L)}(t)\bigr)\bigr]
+
\mathbb{E}_{p(Z^{(L)}(t))}\Bigl[\mathcal{D}_{KL}\Bigl(p\bigl(Y \mid Z^{(L)}(t)\bigr)\big\|q\bigl(Y \mid Z^{(L)}(t)\bigr)\Bigr)\Bigr] \nonumber\\
&\ge
\mathbb{E}\bigl[\log q\bigl(Y \mid Z^{(L)}(t)\bigr)\bigr].
\label{eq:KL_nonneg1}
\end{align}
Substituting \eqref{eq:KL_nonneg1} into \eqref{eq:MI}, and using $-\mathbb{E}_{p(Y)}[\log p(Y)]=H(Y)\ge 0$ for the discrete label $Y$, we obtain
\begin{align}
-I\bigl(Y;Z^{(L)}(t)\bigr)
&\le
-\mathbb{E}_{p(Y,Z^{(L)}(t))}\Bigl[\log q\bigl(Y \mid Z^{(L)}(t)\bigr)\Bigr] - H(Y) \nonumber\\
&\le
-\mathbb{E}_{p(Y,Z^{(L)}(t))}\Bigl[\log q\bigl(Y \mid Z^{(L)}(t)\bigr)\Bigr].
\label{eq:standard up bound1}
\end{align}
Based on the temporal local-dependence assumption, we use the empirical distribution 
$p(Y, Z^{(L)}(t)) \approx \frac{1}{N} \sum_{i=1}^{N} \delta_{(Y_i, Z^{(L)}_i(t))}(Y, Z^{(L)}(t))$~\cite{alemi_deep_2019} to estimate the upper bound \eqref{eq:standard up bound1} as:
\begin{equation}
-\mathbb{E}_{p(Y,Z^{(L)}(t))}\Bigl[\log q\bigl(Y \mid Z^{(L)}(t)\bigr)\Bigr]
\approx -\frac{1}{N} \sum_{i=1}^{N} \log q\bigl(Y_i \mid Z^{(L)}_i(t)\bigr).
\label{eq:empirical}
\end{equation}

Secondly, we prove \eqref{eq:upper_bound2_restate}. Since $Z^{(L)}(t)$ is one of the variables in $S^{(L)}(t)$, the chain rule of mutual information and its non-negativity give
\begin{equation}
I\bigl(Z^{(L)}(t);G([t-\Delta t,t])\bigr)
\le
I\bigl(S^{(L)}(t);G([t-\Delta t,t])\bigr).
\label{eq:tgib_dpi}
\end{equation}
By the chain rule of mutual information over the layers,
\begin{align}
I\bigl(S^{(L)}(t);G([t-\Delta t,t])\bigr)
=\sum_{l=1}^{L}\Bigl[\,
&I\bigl(E^{(l)}(t);G([t-\Delta t,t])\mid S^{(l-1)}(t)\bigr) \nonumber\\
&+
I\bigl(Z^{(l)}(t);G([t-\Delta t,t])\mid S^{(l-1)}(t),E^{(l)}(t)\bigr)\Bigr].
\label{eq:sum_of_two}
\end{align}
For the first term in \eqref{eq:sum_of_two}, we introduce a variational distribution $q\bigl(E^{(l)}(t)\bigr)$. Writing $G$ for $G([t-\Delta t,t])$ and $S$ for $S^{(l-1)}(t)$,
\begin{align}
I\bigl(E^{(l)}(t);G\mid S\bigr)
&=
\mathbb{E}\Bigl[\log\frac{p\bigl(E^{(l)}(t)\mid G,S\bigr)}{p\bigl(E^{(l)}(t)\mid S\bigr)}\Bigr]
\nonumber\\
&=
\mathbb{E}\Bigl[\log\frac{p\bigl(E^{(l)}(t)\mid G,S\bigr)}{q\bigl(E^{(l)}(t)\bigr)}\Bigr]
-
\mathbb{E}_{p(S)}\Bigl[\mathcal{D}_{KL}\Bigl(p\bigl(E^{(l)}(t)\mid S\bigr)\big\|q\bigl(E^{(l)}(t)\bigr)\Bigr)\Bigr]
\nonumber\\
&\le
\mathbb{E}\Bigl[\log\frac{p\bigl(E^{(l)}(t)\mid G,S\bigr)}{q\bigl(E^{(l)}(t)\bigr)}\Bigr]
=\mathrm{EIB}^{(l)}.
\label{eq:eib conclusion}
\end{align}
In the same way, with the variational distribution $q\bigl(Z^{(l)}(t)\bigr)$ and conditioning on $S$ and $E^{(l)}(t)$,
\begin{equation}
I\bigl(Z^{(l)}(t);G\mid S,E^{(l)}(t)\bigr)
\le
\mathbb{E}\Bigl[\log\frac{p\bigl(Z^{(l)}(t)\mid G,S,E^{(l)}(t)\bigr)}{q\bigl(Z^{(l)}(t)\bigr)}\Bigr]
=\mathrm{XIB}^{(l)}.
\label{eq:xib conclusion}
\end{equation}
Finally, substituting \eqref{eq:eib conclusion} and \eqref{eq:xib conclusion} into \eqref{eq:sum_of_two} and incorporating \eqref{eq:tgib_dpi}, we obtain
\begin{equation}
I\bigl(Z^{(L)}(t);G([t-\Delta t,t])\bigr) 
\le
\sum_{l=1}^{L} \mathrm{EIB}^{(l)}+ \sum_{l=1}^{L} \mathrm{XIB}^{(l)}.
\end{equation}
\end{proof}

\section{Structure Enhancer}\label{appendix:structure enhancer}
We analyse the probability that the structure enhancer samples at least one ground-truth neighbour of a node $j$ at time $t$. Let $\hat{E}([t-\Delta t,t])$ be the ground-truth edge set and $\hat{N}_j([t-\Delta t,t])$ the neighbours of $j$ in the ground-truth graph within the time window $[t-\Delta t,t]$, and let $N^{(l)}_j(t)$ be the current $l$-hop neighbours of $j$ at time $t$. As in Section~\ref{sec:sampling_size}, we idealise the $k$ sampling iterations as independent draws: each iteration is, with probability $\alpha\in[0,1]$, a random draw, uniform over $V(t)$, and otherwise a hop-based draw, uniform over $N^{(l)}_j(t)$ at hop $l$ with probability $\beta_l$, where $\sum_{l=1}^{h}\beta_l=1$.

\begin{proposition}[Hit Probability of the Structure Enhancer]\label{prop:enhancer}
Under the sampling model above, for any node $j$ at time $t$,
\begin{equation}
P\bigl(\exists i \in S_j(t): (i,j) \in \hat{E}([t-\Delta t, t])\bigr) \geq 1 - \exp\bigl(-c(t)k\bigr),
\end{equation}
where $S_j(t)$ is the sampled neighbour set of $j$, $k$ is the number of sampling iterations and
\begin{equation}
c(t) = \alpha \frac{|\hat{N}_j([t-\Delta t, t])|}{|V(t)|} + (1-\alpha)\sum_{l=1}^h \beta_l \frac{|\hat{N}_j([t-\Delta t, t])\cap N_j^{(l)}(t)|}{|N_j^{(l)}(t)|},
\label{eq:ct}
\end{equation}
a term of the sum being zero when $N_j^{(l)}(t)$ is empty.
\end{proposition}

\begin{proof}
Let $X_m$ indicate that the $m$-th iteration samples a ground-truth neighbour of $j$, that is, a node $i$ with $(i,j)\in\hat{E}([t-\Delta t,t])$. A random draw does so with probability $|\hat{N}_j([t-\Delta t,t])|/|V(t)|$, and a hop-based draw at hop $l$ with probability $|\hat{N}_j([t-\Delta t,t])\cap N^{(l)}_j(t)|/|N^{(l)}_j(t)|$, so that $P(X_m=1)=c(t)$. With $Y=\sum_{m=1}^{k}X_m$ and the iterations independent,
\begin{equation}
P(Y = 0) = \bigl(1-c(t)\bigr)^k \leq \exp\bigl(-c(t)k\bigr),
\end{equation}
since $1-x\le e^{-x}$. Therefore $P\bigl(\exists i \in S_j(t): (i,j) \in \hat{E}([t-\Delta t, t])\bigr) = 1 - P(Y = 0) \geq 1 - \exp\bigl(-c(t)k\bigr)$.
\end{proof}

\begin{corollary}[Sampling Threshold]\label{cor:threshold}
Let $\delta\in(0,1)$. If $c(t)>0$ and $k \geq \ln(1/\delta)/c(t)$, the enhancer samples a ground-truth neighbour of $j$ within the time window $[t-\Delta t, t]$ with probability at least $1-\delta$.
\end{corollary}
\begin{proof}
By Proposition~\ref{prop:enhancer}, $1-\exp(-c(t)k)\ge 1-\delta$ whenever $k\ge\ln(1/\delta)/c(t)$.
\end{proof}

\begin{corollary}[Uniform Sampling Threshold]\label{cor:threshold_non_degenerate}
Let $\delta\in(0,1)$. If there exists $\varepsilon>0$ with $c(t)\ge\varepsilon$ for all $t$, then $k\ge\ln(1/\delta)/\varepsilon$ iterations give, at every $t$, a probability of at least $1-\delta$ of sampling a ground-truth neighbour of $j$. The required $k$ grows only logarithmically in $1/\delta$ and does not depend on $|V(t)|$, so that the enhancer does not need to sample the entire graph.
\end{corollary}
\begin{proof}
Immediate from Corollary~\ref{cor:threshold}, since $\ln(1/\delta)/c(t)\le\ln(1/\delta)/\varepsilon$.
\end{proof}

The random component alone gives $c(t)\ge\alpha|\hat{N}_j([t-\Delta t,t])|/|V(t)|$, which vanishes on a sparse growing graph, where the number of nodes grows faster than the neighbourhood of a node. On such a graph, a lower bound $\varepsilon$ must come from the hop-based component, which is large when the ground-truth neighbours of $j$ lie near the nodes it is already connected to; for a node without neighbours, random sampling first establishes these connections. The two steps of the enhancer serve these two roles. The proposition bounds the probability of sampling a ground-truth neighbour; the effect of the sample size on prediction is evaluated in Section~\ref{sec:sampling_size}.

\section{Datasets}\label{appendix:dataset}
\begin{table}[htb]
\centering
\caption{Summary of Dataset Statistics}
\label{tab:datasets}
\setlength{\tabcolsep}{4pt}
\footnotesize
\begin{tabular}{@{}lrrrrrrr@{}}
\toprule
\textbf{Dataset} & \textbf{Nodes} & \textbf{Edges} & \textbf{Edge Feature} & \textbf{Node Feature} & \textbf{Time} & \textbf{Duration} & \textbf{Avg.} \\
& & & \textbf{Dimension} & \textbf{Dimension} & \textbf{Granularity} & & \textbf{Degree} \\
\midrule
Wikipedia     & 9,227     & 157,474   & 172 & 172 & Unix timestamp & 1 month   & 34.14 \\
MOOC          & 7,145     & 411,749   & 4   & 0   & Unix timestamp & 17 months & 115.15 \\
UCI           & 1,899     & 59,835    & 0   & 0   & Unix timestamp & 196 days  & 63.00 \\
Social Evo.   & 74        & 2,099,519 & 0   & 2   & Unix timestamp & 8 months  & 56,743.76 \\
\bottomrule
\end{tabular}
\end{table}

\paragraph{Wikipedia}
This dataset chronicles user editing activities on Wikipedia pages through a bipartite interaction graph structure. The network comprises users and pages as nodes, while the links represent editing behaviors with their corresponding timestamps. Each interaction is characterized by a 172-dimensional Linguistic Inquiry and Word Count (LIWC) feature vector. 

\paragraph{MOOC}
The MOOC dataset presents a bipartite interaction network focusing on online learning behavior. The nodes represent students and course content units, including videos and problem sets. Each interaction link is annotated with a 4-dimensional feature vector that characterizes the access behavior of students to specific content units. 

\paragraph{UCI}
This dataset captures communications within a Facebook-like online network among students at the University of California at Irvine. The temporal network records interactions with second-level granularity, providing a detailed view of communication patterns. 

\paragraph{Social Evolution}
This dataset represents a mobile phone proximity network that monitors daily activities within an undergraduate dormitory. The network exhibits a remarkably high average degree, significantly surpassing that of other datasets. In our study, we utilize a two-week subset of this dataset, reducing the average degree from 56743.76 to 1,013.59. This presents a challenging scenario for inductive learning due to its compressed temporal window.

\section{Detailed Descriptions of Baseline Models}\label{appendix:baselines}

\paragraph{JODIE}
JODIE employs a RNN architecture to model temporal dynamics in networks. The model maintains temporal embeddings for each node, which are updated through the RNN whenever an interaction occurs. These embeddings incorporate both the historical interaction patterns and the time elapsed since the last interaction. The temporal evolution of node embeddings is achieved through a linear projection mechanism that accounts for the time difference between interactions.

\paragraph{TGAT}
TGAT utilizes a self-attention mechanism to simultaneously capture spatial and temporal information. The model employs a novel time encoding scheme that combines node features with temporal information through trainable parameters. The self-attention mechanism aggregates information from temporal neighborhoods, enabling the model to learn time-aware node representations. The final predictions are computed using a multi-layer perceptron that processes paired node representations.

\paragraph{TGN}
TGN presents a hybrid approach combining RNN and self-attention mechanisms. The model's core component is a memory module that maintains and updates node states, which serve as compressed representations of historical interactions. The memory updater, implemented as an RNN, processes new interactions to modify node states. Node embeddings are generated by aggregating information from K-hop temporal neighborhoods using self-attention mechanisms.

\paragraph{DyRep}
DyRep adopts an RNN-based architecture for dynamic graph representation learning. The model features a state update mechanism that processes each interaction as it occurs. A distinguishing characteristic is its temporal-attentive aggregation module, which effectively incorporates evolving structural information in dynamic graphs. This approach enables the model to capture both topological and temporal dependencies in the network.

\paragraph{DyGFormer}
DyGFormer implements a self-attention-based approach with a unique temporal processing strategy. Rather than processing interactions individually, the model segments each node's interaction sequence into patches, which are then processed by a transformer architecture. This patch-based processing enables efficient handling of temporal information and allows the model to capture long-range dependencies in the interaction sequences.

\paragraph{GraphMixer}
GraphMixer presents a simplified architecture based on multi-layer perceptrons (MLPs). The model employs a deterministic time encoding function in conjunction with an MLP-Mixer based link encoder to process temporal information. Node representations are computed through a straightforward neighbor mean-pooling mechanism, resulting in an efficient yet effective approach to temporal graph learning.

\paragraph{CAW}
CAW combines RNN and self-attention mechanisms in a novel way. The model utilizes temporal random walks to extract network motifs, representing network dynamics through continuous-time aware walks. These walks are encoded using RNNs and the resulting representations are aggregated using self-attention mechanisms. A key innovation is the replacement of node identities with hitting counts based on sampled walks, enabling effective correlation between motifs.

\section{Training Time Analysis}\label{appendix:train time}
\subsection{Time Complexity}
\begin{figure}[htb]
    \centering
    \includegraphics[width=0.6\linewidth]{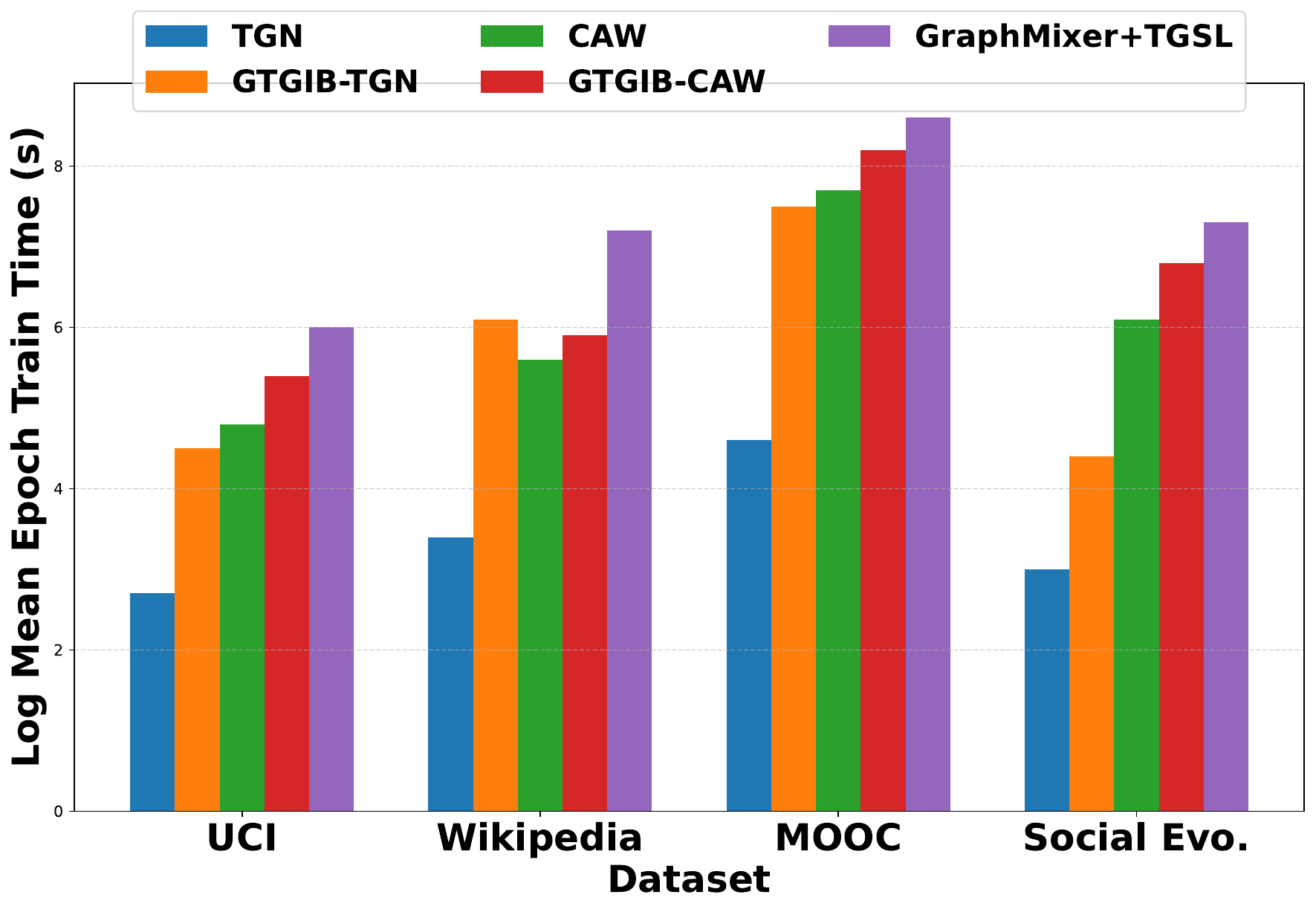}
    \caption{Training time comparison. Although GTGIB introduces additional time complexity, it remains less time-consuming than other GSL algorithms, particularly GraphMixer+TGSL. For the single-layer CAW model, the primary computational cost stems from the original model’s random walk mechanism, resulting in only a marginal increase in training time due to the GTGIB framework.}
    \label{fig:train_time}
\end{figure}

In this section, we analyze the time complexity of TGN, GTGIB-TGN, CAW, and GTGIB-CAW.

\paragraph{\textbf{Structure Enhancer.} }
The random sampling enhancer has a time complexity of $\mathcal{O}(k|E|)$, while hop-based sampling requires $\mathcal{O}(hk|E|)$, where $k$ is the number of sampled nodes, $h$ is the maximum hop and $|E|$ denotes the total number of edges. Hence, the structure enhancer exhibits a complexity of $\mathcal{O}(hk|E|)$.

\paragraph{\textbf{TGIB Filtering.} }
The single-layer TGIB filtering module has a time complexity of $\mathcal{O}(d|E|)$, where $d$ represents the embedding dimension.

\paragraph{\textbf{Overall Complexity.} }

\textbf{TGN:} $\mathcal{O}\left(L(\bar{n} + d)d |E|\right)$, where $\bar{n}$ is the average node degree and $L$ is the number of layers.

\textbf{CAW:} $\mathcal{O}(wN|V|)$, where $w$ is the random walk length and $N$ is the number of random walks per node.

\textbf{GTGIB-TGN:} $\mathcal{O}\left(L\bar{\pi} (\bar{n} + d)d|E|+(Ld + hk)|E|\right)$, where $\bar{\pi}$ represents the average edge retention rate after TGIB filtering. As observed, the complexity of the base TGN model remains largely unchanged, with the additional cost primarily stemming from the per-layer application of TGIB, which contributes an additional $\mathcal{O}(d|E|)$.

\textbf{GTGIB-CAW:} $\mathcal{O}\left(wN|V|+(d + hk)|E|\right)$. The dominant computational cost still arises from the random walk mechanism of CAW, with the added TGIB overhead being relatively minor.

Combining these theoretical results with the empirical findings in Figure~\ref{fig:train_time}, we observe that the training time of GTGIB-TGN increases, due to the per-layer integration of TGIB. However, its overall computational cost remains lower than that of CAW and GraphMixer+TGSL. 

On the other hand, the additional computational cost introduced by GTGIB-CAW is relatively minor compared to CAW itself, indicating that the random walk mechanism still dominates the overall complexity. Additionally, its training time remains lower than that of GraphMixer+TGSL.

\subsection{Time Consumption Distribution}
\begin{table}[htbp]
\centering
\caption{Runtime and relative overhead ($\Delta$\%) of the CAW backbone and various GTGIB-CAW configurations on an NVIDIA A100 GPU on UCI dataset. The $\Delta$\% column indicates the fraction of runtime attributable to the extra GTGIB-CAW module.}
\label{tab:gtgib-caw-runtime}
\begin{tabular}{lcc}
\toprule
Model & Runtime & $\Delta$\% \\
\midrule
CAW (backbone)               & 51.3  & 65.33\% \\
GTGIB-CAW (rand only)        & 66.2  & 19.28\% \\
GTGIB-CAW (hop only)         & 61.5  & 13.20\% \\
GTGIB-CAW (TGIB only)        & 56.5  &  6.73\% \\
GTGIB-CAW (rand \& hop)      & 69.0  & 22.90\% \\
GTGIB-CAW (all)              & 77.3  & 33.64\% \\
\bottomrule
\end{tabular}
\end{table}
We cannot run a single part of the model, but we analyze its training time via ablation. Table~\ref{tab:gtgib-caw-runtime} demonstrates that the random walk module, CAW, accounts for a large portion of the complexity. From the table, the structure enhancer consumes a significant portion of runtime while TGIB remains highly efficient. Notably, in GSL-based methods, the structure enhancer is relatively efficient, as confirmed by Table~\ref{tab:epoch-times}.

\begin{table}[htbp]
\centering
\caption{Mean epoch times (in seconds) on four datasets using one NVIDIA V100 GPU. Comparison of the CAW backbone, GTGIB-CAW, and baselines (GraphMixer backbone and GraphMixer+TGSL). $\Delta$\% indicates the runtime increase.}
\label{tab:epoch-times}
\begin{tabular}{lcccccc}
\toprule
Dataset        & CAW     & GTGIB-CAW & $\Delta$\%     & GraphMixer & GraphMixer+TGSL & $\Delta$\%      \\
\midrule
Wikipedia      & 290.9   & 382.9     & 31.64\% &  71.0      & 1407.7          & 1882.71\% \\
UCI            & 140.0   & 210.47    & 50.34\% &  11.3      &  398.0          & 3422.12\% \\
MOOC           & 2213.5  & 3508.00   & 58.48\% & 186.7      & 5194.30         & 2682.16\% \\
Social Evo.    & 425.0   & 691.25    & 62.65\% &  37.0      & 1650.0          & 4359.46\% \\
\bottomrule
\end{tabular}
\end{table}
 Although CAW is slower than GraphMixer due to its random walk mechanism, GTGIB-CAW’s mean runtime increases by only about 50.78\% compared to CAW, two orders of magnitude lower than the 3086.61\% of GraphMixer+TGSL. CAW accounts for 66.77\% mean runtime in GTGIB-CAW, much larger than 3.43\% GraphMixer of GraphMixer+TGSL. This efficiency stems from our 2-step GSL-based structure enhancer, which has much lower computational complexity than GraphMixer+TGSL’s parameterized structure learner and contrastive learning. 
 
\end{document}